\documentclass[fleqn,10pt]{wlscirep}
\usepackage[utf8]{inputenc}
\usepackage[T1]{fontenc}

\usepackage[utf8]{inputenc} 
\usepackage[T1]{fontenc}    
\usepackage{hyperref}       
\usepackage{url}            
\usepackage{booktabs}       
\usepackage{amsfonts}       
\usepackage{nicefrac}       
\usepackage{microtype}      
\usepackage{xcolor}         

\usepackage{tabularx}

\usepackage{times}
\usepackage{soul}
\usepackage{url}
\usepackage{graphicx}
\usepackage{amsmath}
\usepackage{amssymb}
\usepackage{amsthm}
\usepackage{booktabs}
\usepackage{algorithm}
\usepackage[noend]{algpseudocode}
\usepackage[switch]{lineno}
\usepackage{todonotes}
\usepackage{tabularray}
\UseTblrLibrary{diagbox}

\title{Generative inpainting of incomplete Euclidean distance matrices of trajectories generated by a fractional Brownian motion}

\author[1]{Alexander Lobashev}
\author[1]{Dmitry Guskov}
\author[1,2,*]{Kirill Polovnikov}
\affil[1]{Skolkovo Institute of Science and Technology, Moscow, Russia}
\affil[2]{Laboratory of Complex Networks, Center for Neurophysics and Neuromorphic Technologies}

\affil[*]{kipolovnikov@gmail.com}

\begin{abstract}

Fractional Brownian motion (fBm) features both randomness and strong scale-free correlations, challenging generative models to reproduce the intrinsic memory characterizing the underlying stochastic process. Here we examine a zoo of diffusion-based inpainting methods on a specific dataset of corrupted images, which represent incomplete Euclidean distance matrices (EDMs) of fBm at various memory exponents $H$. Our dataset implies uniqueness of the data imputation in the regime of low missing ratio, where the remaining partial graph is rigid, providing the ground truth for the inpainting. We find that the conditional diffusion generation readily reproduces the built-in correlations of fBm paths in different memory regimes (i.e., for sub-, Brownian and super-diffusion trajectories), providing a robust tool for the statistical imputation at high missing ratio. Furthermore, while diffusion models have been recently shown to memorize samples from the training database, we demonstrate that diffusion behaves qualitatively different from the database search and thus generalize rather than memorize the training dataset. As a biological application, we apply our fBm-trained diffusion model for the imputation of microscopy-derived distance matrices of chromosomal segments (FISH data) -- incomplete due to experimental imperfections -- and demonstrate its superiority over the standard approaches used in bioinformatics.
\end{abstract}
\begin{document}

\flushbottom
\maketitle

\thispagestyle{empty}

\section*{Introduction}

Diffusion probabilistic models are gaining popularity in the field of generative machine learning due to their ability to synthesize diverse and high-quality images from the training distribution. The iterative denoising approach taken by diffusion \cite{ho2020denoising,song2020score,Sohl-Dickstein} outperforms in quality of generated samples the previously used schemes \cite{Dhariwal}, such as VAEs \cite{vahdat2020nvae,Diederik} and GANs \cite{karras2020analyzing,Brock,Goodfellow}, and has demonstrated a distinctive potential in scalability \cite{Ramesh}. 
Recently, several conditional diffusion-based generation methods have been developed \cite{lugmayr2022repaint,wang2022zero,kawar2022denoising}, allowing for effective inpainting of masked images using the pre-trained unconditional diffusion model. Still, whether the diffusion-based inpainting can learn and reproduce the intrinsic non-local dependencies in the pixels of the image drawn from a particular statistical ensemble has remained unaddressed.
Furtheremore, recent studies by \cite{carlini2023extracting,somepalli2023diffusion} 
suggest that modern text-to-image generative diffusion models, such as Dalle-2 \cite{Ramesh}, Imagen \cite{saharia2022photorealistic}, or StableDiffusion \cite{rombach2022high}, tend to recall samples from their training databases, raising questions about their generalization capabilities and bringing up the copyright infringement concerns during the diffusion training process.

In this paper we consider a dataset of incomplete EDMs and propose to approach the EDM completion problem as the image inpainting via conditioning of the diffusion generative models. Importantly, the possibility of existence of the ground truth of the inpainting in the EDM dataset uniquely allows one to evaluate the quality of the conditional generation at the instance level. At high missing ratio, however, the solution of EDM completion does not exist and one has to rely on the ensemble-level metrics such as Fréchet Inception Distance (FID). Here we ask: can the diffusion model learn the intrinsic correlations between the entries of the matrix when an ensemble of such matrices is given and statistically reproduce them upon the inpainting? In order to explore the modern generative models at this novel angle we consider the pairwise distances between the points of a discrete fractional Brownian process (fBm), the simplest Gaussian generalization of a Brownian motion with strong scale-free correlations. The built-in memory in the fBm process can induce a non-Brownian exponent of the second moment (also known as the mean-squared displacement of a particle undergoing the anomalous diffusion, \cite{Metzler00}), which is translated into strong couplings between the pixels in the distance matrix.

Imputation of missing data has recently got a second wind with the development of high-throughput experimental techniques in chromosome biology. Diffusion models have been recently applied to generate and enhance protein and DNA datasets \cite{watson2023novo,ingraham2023illuminating,wang2023hicdiff}. Hi-C and FISH experiments have provided significant new insights into the fractal 
(non-Brownian) folding of chromosomes \cite{Lieberman09,Bintu18}, despite the data being noisy and incomplete \cite{imakaev,galitsyna}. In particular, we and others have recently shown that the spatial organization of human chromosomes without loop-extruding complexes (cohesin motors) statistically resembles the ensemble of fractal trajectories with the fractal dimension $d_f=3$ \cite{Polovnikov23,Polovnikov18,Polovnikov19,polovnikovprl18,TammPolovnikov,Lieberman09}. Such an ensemble -- leaving aside the biophysical principles of such organization -- can be modelled as trajectories of a subdiffusive fBm particle with $H=1/3$ \cite{Polovnikov18}. This suggests an important statistical insight for the downstream data analysis \cite{polovnikov20,ulianov21} (also relevant for the worm connectome datasets \cite{polovnikov23sr}). 

FISH imaging experiments produce datasets that represent matrices of pairwise distances between chromosomal loci on single cells, which are obtained in multiplex microscopy. Thus each matrix corresponds to internal distances within a given chromosomal segment in a given cell. Occasionally, some data in the matrices is masked due to experimental imperfections (biochemistry of the protocol) posing a real challenge for the methods of the downstream analysis. In particular, inference of features of the 3D organization at the single cell level is notoriously obscured by the sparsity of the dataset at hand \cite{galitsyna}. Here we for the first time propose to use the modern generative AI for the inpainting of missing values and completion of experimentally-derived FISH matrices. For this aim we deploy the pre-trained fBm diffusion benchmark at $H=1/3$, thus virtually taking into account the intrinsic correlations present in the fractal chromosome trajectories \cite{Polovnikov18,Polovnikov23}.

The structure of this paper is as follows. In section Background we formulate the EDM completion problem as the image inpainting task, discuss some of classical results from discrete mathematics related to the existence and uniqueness of the EDM completion. In the following section (a) we demonstrate that the unconditional diffusion generation can learn and reproduce non-local correlations in the images (i.e., matrices) representing EDMs of fBm at various memory exponents $H$: for subdiffusion ($H<1/2$), normal ($H=1/2$) and superdiffusion ($H>1/2$) in the single framework; (b) we apply the diffusion-based inpainting for the EDM completion problem showing that it results in low-rank solutions with the proper fBm-like statistics. We further demonstrate that the diffusion generation is qualitatively different from the database search, regardless of the database size, which is being in contrast with the most recent studies \cite{carlini2023extracting,somepalli2023diffusion}. Finally, in the last section, we illustrate how the pre-trained fBm diffusion model can be applied for the imputation of missing values in the chromosomal distance matrices derived from FISH experiments. We demonstrate superior performance of the diffusion-based inpaiting as compared to classical bioinformatics approaches. Our results pave the way for an accurate quantification of the cell-to-cell variability in the genome folding and, broadly, showcase the importance of generative AI in the omics data analysis.

\section*{Background}

\subsection*{Euclidean distance matrices}
In this paper we deal with $n \times n$ matrices $A$ of squares of pairwise distances between $n$ points $x_1, x_2, ..., x_n$ in the $D$-dimensional 
Euclidean space. For the purposes of this paper, we considered the case of $D=3$. Such matrices $A=\{a_{ij}\}$ satisfying
\begin{equation}
a_{ij} = ||x_i - x_j||^2, \quad x_i \in \mathbb{R}^D
\label{1}
\end{equation}
are called Euclidean distance matrices (EDM). Clearly, $A$ is a symmetric ($a_{ij}=a_{ji}$) and hollow ($a_{ii}=0$) matrix with non-negative values, $a_{ij} \ge 0$. As a distance matrix in Euclidean space it further satisfies the triangle inequality, $\sqrt{a_{ij}} \le \sqrt{a_{in}} + \sqrt{a_{nj}}$. The latter constraint imposes essential non-linear relationships between the entries of an EDM, making its rank $r$ independent of $n$ for sufficiently large amount of points $n$ in general position, i.e. $r = \min(n, D+2)$.

Any uncorrupted (complete, noise-less and labelled \cite{Dokmanic15}) EDM $A$ allows for the unique reconstruction of the original coordinates $\{x_i\}$ up to 
rigid transformations (translations, rotations and reflections). Such reconstructions are called realizations of $A$. Due to the straightforward relation between an EDM (see Eq. \ref{1}) and the corresponding Gram matrix $g_{ij}=x^T_i x_j$,
\begin{equation}
a_{ij} = g_{ii} - 2g_{ij} + g_{jj},
\label{2}
\end{equation}
a realization of the distance matrix $a_{ij}$ consists of the origin ($x_1=0$) and the principal square root of the $(n-1)\times (n-1)$ matrix $\tilde{g}_{ij}$
\begin{equation}
\tilde{g}_{ij} = \frac{1}{2} (a_{1i} - a_{ij} + a_{1j}),
\label{3}
\end{equation}
provided that $\tilde{g}_{ij}$ is positive semidefinite with rank $D$ (for points in general position). The latter is known as the Schoenberg criterion \cite{Schoenberg35,Young38}. Other classical conditions for the existence of a realization of the complete EDM make use of the relations involving the Cayley-Menger determinants \cite{Menger31} and allow to decide whether this realization exists in the given space dimension $D$. Note that the general rank property of EDMs outlined above follows simply from Eq. 2: since the rank of Gram matrix $g_{ij}$ is $D$ and the ranks of the other two terms in the equation is $1$, the rank of $a_{ij}$ cannot exceed $D+2$. For other interesting properties of EDMs we refer the reader to classical textbooks on the topic \cite{Krislock12,Mucherino12,Liberti14}. 

Noisy measurements of pairwise distances notably violate the properties of EDMs discussed above. In this case one is interested in the \textit{optimal} embedding of the points in the space of desired dimension. For that low-rank approximations by means of SVD or EVD of the Gram-like matrix $\tilde{g}_{ij}$ from Eq. \ref{3} are typically implemented in the spirit of the classical multidimensional scaling approach \cite{Mead92,Dokmanic15}.

\subsection*{Reconstruction of incomplete EDMs}

A case of incomplete distance matrix, where a particular set of pairwise distances in Eq. 1 is unknown, is a prominent setting of EDM corruption that we study in this paper. 

\textbf{EDM completion problem}. \textit{Let us specify $m<\binom{n}{2}$ missing pairwise distances between $n$ points $x_1, x_2, ..., x_n$ of the D-dimensional Euclidean space by means of the symmetric mask matrix $B = \{b_{ij}\}$ consisting of $2m+n$ zeros and $n^2-n-2m$ ones. By definition, $b_{ij}=0$ if the distance between $i$ and $j$ is unknown or $i=j$, and $b_{ij}=1$ otherwise. The matrix $B$ is the adjacency matrix of the resulting partial graph. That is, we have a matrix $\tilde{A}=\{\tilde{a}_{ij}\}$:
\begin{equation}
\tilde{a}_{ij} = a_{ij} b_{ij}^{-1} \quad \text{for}\quad b_{ij}=1,
\label{tila}
\end{equation}
while $\tilde{a}_{ij}$ is undefined where $b_{ij}=0$. The goal is, given an incomplete matrix $\tilde{A}$, defined by Eq. \ref{tila}, restore $\binom{n}{2}-m$ missing pairwise distances, while preserving the known $m$ distances. For approximate EDM completions, one seeks an EDM matrix $A$, such that the following Frobenius norm
\begin{equation}
||B \odot (A-\tilde{A})||_F^2 \to \min
\end{equation}
is minimized. For precise completions, if they exist, this norm simply equals to zero. In some formulations for approximate completions the constraint $b_{ij}=1$ at known distances can be relaxed to $b_{ij}>0$.}

Clearly, for the EDM completion to exist the matrix $\tilde{A}$ must satisfy all the key properties of EDMs over the known entries, such as symmetricity, hollowness, non-negativity, as well as the triangle inequality at the known triples $(i, j, n), \text{s.t.} \; b_{ij}b_{in}b_{jn} = 1$. If these trivial conditions are satisfied, the matrix $\tilde{A}$ is called a partial EDM (i.e. every fully specified principal submatrix of $\tilde{A}$ is itself an EDM). However, it is not sufficient: for example, when the graph $B$ has a long (with length $l \ge 4$) chordless cycle, one can choose the distances along the cycle such that $\tilde{A}$ does not allow for the completion. The classical GJSW theorem \cite{Grone84} states that a partial EDM $\tilde{A}$ allows for the completion if the graph of specified distances $B$ is \textit{chordal}, i.e. it has no holes or cycles of length $l \ge 4$ without chords. Still, the solution can be non-unique.

\begin{figure}[t]
\centering    
\includegraphics[width=0.7\textwidth]{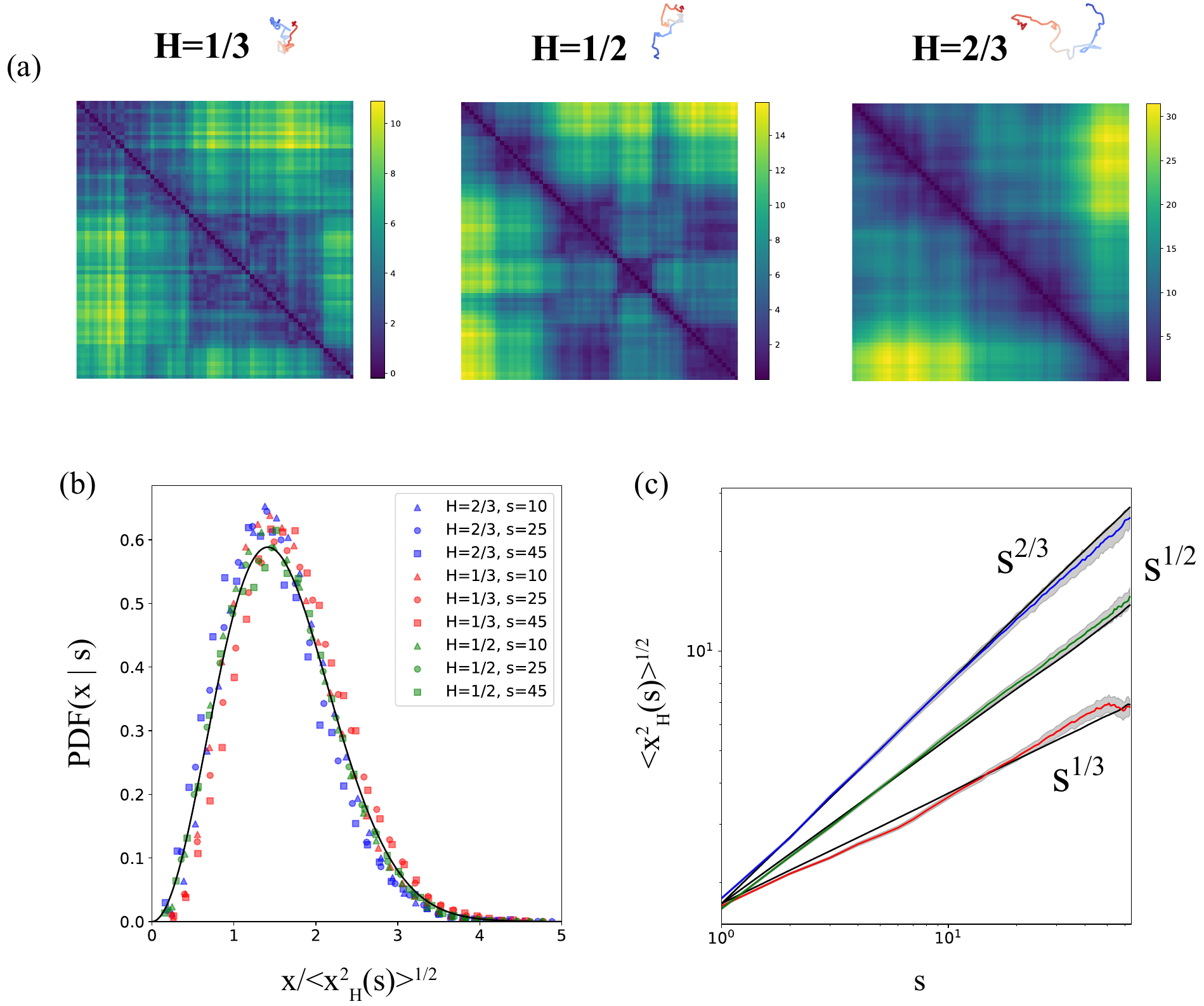}    
\caption{(a) Euclidean distance matrices and the corresponding fBm trajectories, generated by the unconditional diffusion model, for the three values of Hurst exponent: $H=1/3$ (subdiffusion),  $H=1/2$ (normal diffusion) and $H=2/3$ (superdiffusion). The trajectories were obtained using gradient optimization of three-dimensional coordinates to match the generated distance matrices. The color changes from red to blue along the trajectory. (b) Collapsed probability densities of the diffusion-generated pairwise distances between two points on the trajectory separated by contour distance $s$. The black curve corresponds to standard Gaussian. (c) Scaling of the typical distances as a function of the contour length $s$ for $H=1/3$ (red), $H=1/2$ (green) and $H=2/3$ (blue). EDM samples from the generated ensemble fill the grey intervals and thick color lines correspond to the ensemble-averaged curves. Black lines correspond to the training databases.}    
\label{fig:figure_01}    
\end{figure}

\subsubsection*{When is the completion unique?}

In this paper we consider a specific case, when the incomplete matrix $\tilde{A}$ is obtained from a full EDM matrix $A$ by masking some ($m$) distances. 
Thus, the solution of EDM completion of $\tilde{A}$ surely exists. 
A key relevant question for us, essential for proper interpretation of the diffusion predictions, is whether the solution of the matrix completion is unique.

In fact, uniqueness of the EDM completion is equivalent to uniqueness of the distance-preserving immersion $\{x_i\}$ of the partial graph $B$ with the lengths of edges $\tilde{A}$ into the metric space $\mathbb{R}^D$. 
The corresponding bar-and-joint framework $(B, \{x_i\})$ is called rigid (universally rigid), if all these configurations (in any space dimension $D$) are equivalent up to distance-preserving transformations \cite{Connelly22,Connelly15}.
There has been a series of classical results addressing sufficient conditions for the rigidity of frameworks through the stress and rigidity matrices (see \cite{Connelly22} for review), as well as by evoking semi-definiteness of $A$ instead of the non-negativity \cite{Alfakih10,Gortler14,Linial95}. Still, in practice testing for rigidity is known to be a NP-hard problem, unless the points are in general position \cite{Saxe79,Connelly15}. 

The more distances $m$ missing in $\tilde{A}$, the more probable the solution of EDM completion is non-unique. Each ensemble of partial EDMs with exactly $2m$ missing entries can be characterized by the following missing ratio  
\begin{equation}
\mu=2m/(n(n-1)). 
\end{equation}
This number reflects the typical amount of constraints per each vertex ($\mu n$) in the partial graph. Though a useful intuitive measure, the missing ratio alone is not telling of the graph rigidity. In particular, in $D$ dimensions \textit{it is not sufficient} to have $D$ constraints per vertex to ensure the graph rigidity (as a counterexample, imagine two cliques of size $>D$ glued at a single vertex).

\subsubsection*{A greedy algorithm to check for the graph rigidity}\label{rigidity}

Here we suggest the following greedy algorithm that checks for the rigidity of a given graph $B$. We describe it for $D=3$, however, it can be simply generalized for an arbitrary $D$.  

The algorithm sequentially chooses and adds vertices one by one to a subgraph, ensuring the growing subgraph at each step remains rigid. The key idea is that the coordinates of a new vertex in $D$ dimensions can be uniquely determined, if it is connected to at least $D+1=4$ vertices of the rigid subgraph. Following this idea, on the initial step (i) the algorithm identifies the maximal clique with not less than $4$ vertices. As the clique has a complete EDM, there is the corresponding unique realization in the metric space (all cliques are rigid). Then, (ii) the algorithm seeks and adds a new external vertex, maximally connected with the rigid subgraph, but having not less than $4$ edges.
This process continues (iii-iv...) until all vertices are included to the subgraph, or none can be further added. If all the vertices are eventually included, the algorithm confirms that the given graph $B$ is rigid and the EDM completion of $\tilde{A}$ is unique. See Fig. S1 for the graphical sketch and Methods for the pseudo-code of the algorithm.

\begin{figure}[t]
\centering    
\includegraphics[width=0.48\textwidth]{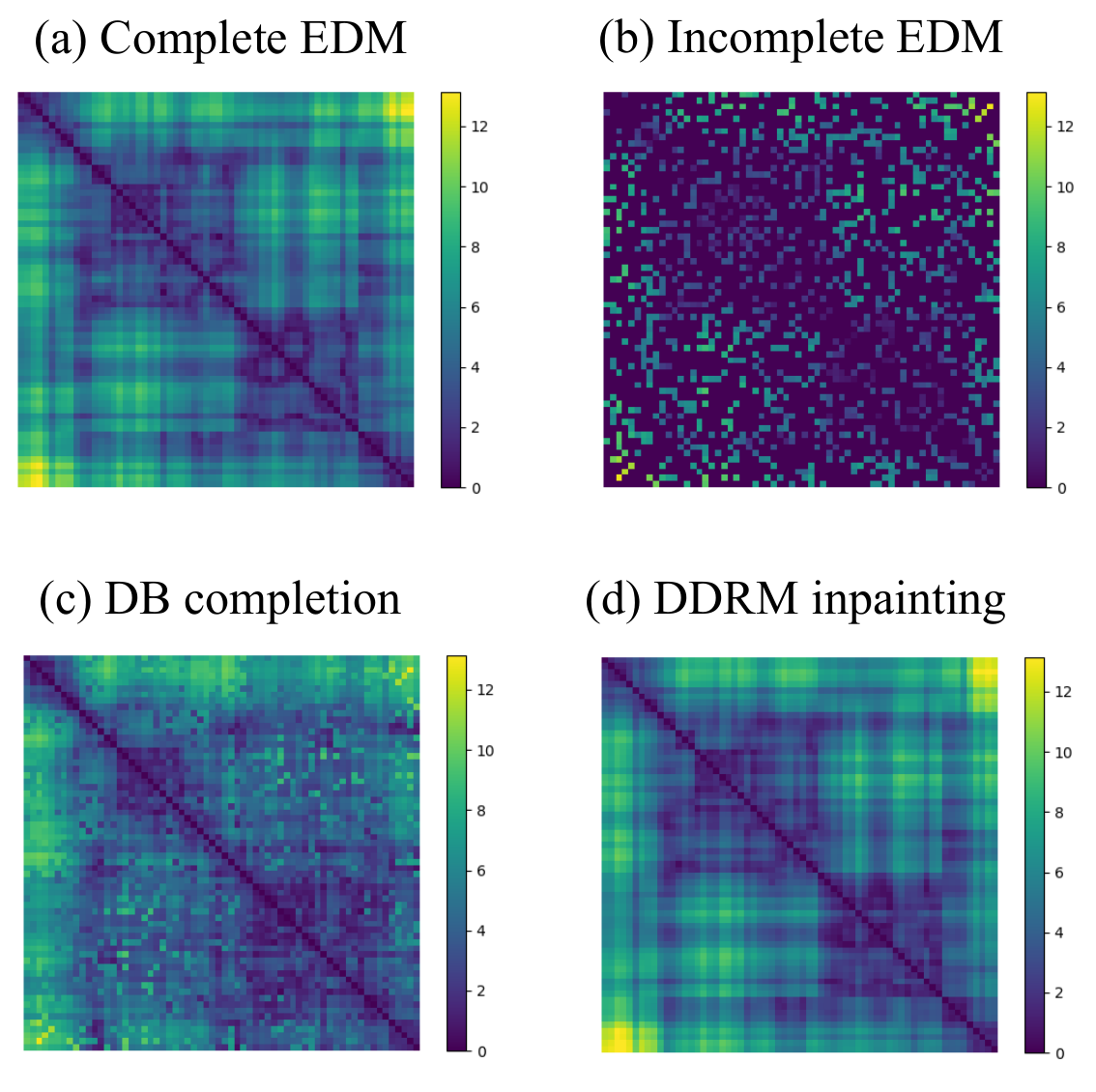}    
\caption{Examples of EDM completion. (a) Complete EDM for a Brownian trajectory ($H=1/2$). (b) A corrupted EDM obtained from (a) by masking distances at sparsity $\mu=0.75$. No exact solution exists at such a high sparsity. (c) The most similar matrix from the database search (the database size is $M=2 \times 10^4$). (d) The EDM completion obtained by the DDRM inpainting.}    
\label{fig:figure_02}    
\end{figure}

\subsection*{Fractional Brownian motion}

Fractional Brownian motion is one the simplest generalizations of Brownian motion that preserves Gaussianity of the process, but introduces strong memory effects \cite{Mandelbrot68}.
By definition, fBm is a Gaussian process $B_H(t)$ on the interval $[0, T]$ that starts at the origin, $B(0)=0$, and has the following first two moments:
\begin{equation}
\begin{split}
&\langle B_H(t) \rangle = 0; \\
&\langle B_H(t)B_H(t') \rangle = \frac{1}{2} \left(t^{2H} + t'^{2H} - |t-t'|^{2H}\right)
\end{split}
\label{fbm}
\end{equation}
and $0<H<1$ is the Hurst parameter (the memory exponent). As it remains Gaussian and ergodic \cite{Metzler00,Metzler14}, the fBm model of anomalous diffusion often allows for analytical treatment. Clearly, the increments of fBm are not independent for $H \ne 1/2$, at which it reduces to Brownian motion. The mean-squared displacement of 
fBm, $\langle B_H^2(t) \rangle$, describing how far the trajectory spreads from the origin at time $t$, can be obtained from Eq. \ref{fbm}:
\begin{equation}
x^2(t) \equiv \langle B_H^2(t) \rangle = t^{2H}.
\label{r2}
\end{equation}
The parameter $H$ physically characterizes fractality ("roughness") of the trajectory.
The autocorrelation of the fBm increments is given by the second derivative of $x^2(t)$: 
\begin{equation}
\langle dB_H(t) dB_H(0) \rangle \sim \frac{2H(2H-1)}{t^{2-2H}}, \quad H = 1/2 \pm \varepsilon
\label{corr}
\end{equation}
Thus, normal diffusion (delta-functional 
correlations) is a particular case of fBm at $H=1/2$, subdiffusion (negative power-law correlations between the increments) corresponds to $H<1/2$, while superdiffusion (positive power-law correlations) corresponds to $H>1/2$. 

In this paper we consider a discrete-time process $\{x_i\}_{i=1}^N$ parameterized by $\{s_i\}_{i=1}^N$ along the trajectory with the fBm statistics, Eq. \ref{fbm} ($x_0=0$ at $s_0=0$). That is, the mean-squared displacement is $\langle x^2_s \rangle = a^2 s^{2H}$, where $a$ is the typical displacement at a single "jump"; the pdf of the end-to-end spatial distance $x=|x_i-x_j|$ between the points $s_i$ and $s_j$ of the trajectory depends only on the contour distance $s=|s_i-s_j|$ and reads
\begin{equation}
P(x | s) = \left(\frac{D}{2\pi a^2 s^{2H}}\right)^{D/2} \exp \left(-\frac{D x^2}{2a^2 s^{2H}}\right)
\label{gauss}
\end{equation}
Anomalous diffusion featuring strong memory effect can have different physical origins; examples include charge transport in semiconductors, cellular and nuclear motion etc. (see \cite{Bouchaud90,Metzler14} for a comprehensive review).
Recently, an fBm model of chromosome folding has been proposed, where fractal chromosome conformations are modelled as trajectories of a subdiffusive fBm particle \cite{Polovnikov18,Polovnikov19,Polovnikov23}. 

In what follows, we train the diffusion model to generate complete EDM images $A=\{a_{ij}\}$ (see Eq. 1) of discrete fBm trajectories $\{x_i\}_{i=1}^N$ for different values of the Hurst exponent. The essential new competence of the diffusion generative model that we test is its ability to learn the power-law correlations in the fBm trajectories. This is crucial for the proper inpainting of the missing data in the distance matrix.

\section*{Diffusion-based generation of fBm trajectories}
\subsection*{Unconditional generation}

\begin{figure}
\centering  
\includegraphics[width=\textwidth]{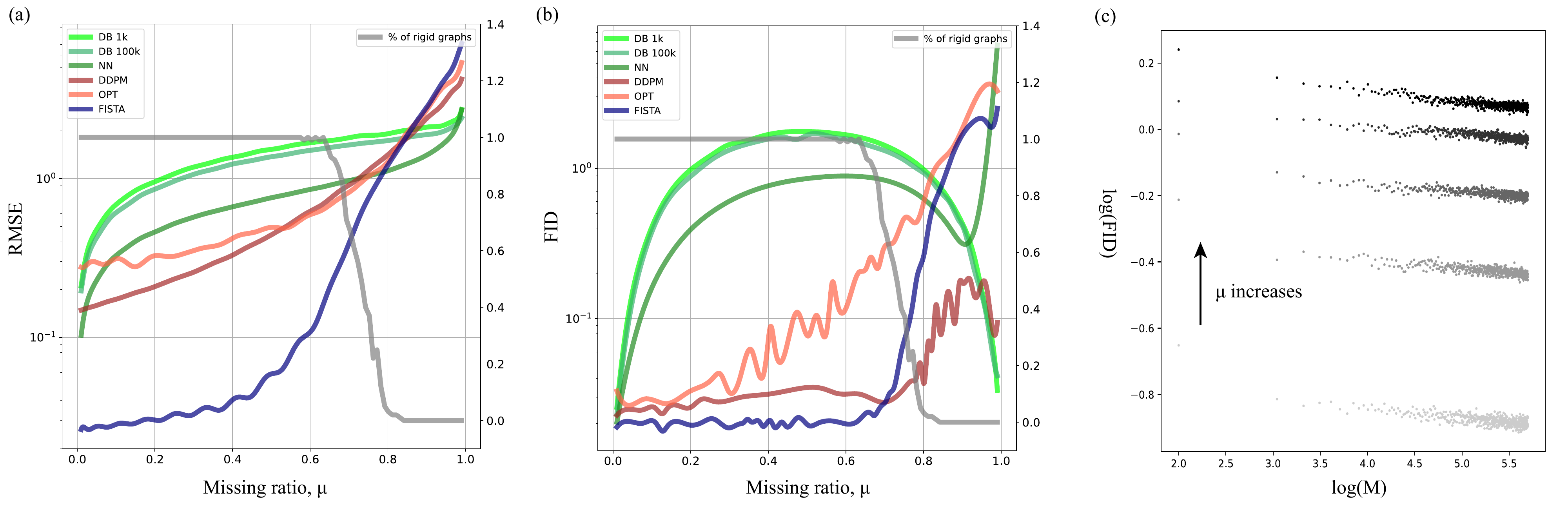}  
\caption{RMSE (a) and FID (b) plots as a function of missing ratio $\mu$ for different data imputation methods ($H=1/2$). The fraction of rigid graphs is shown in the second axis (grey). (c) Log of FID of the matrices reconstructed via the database search as a function of the log of the database size $M$. Scatters for different values of sparsity $\mu=0.05, 0.1, 0.15, 0.19, 0.24$ are shown, corresponding to the regime where the unique EDM completion exists.}  
\label{fig:my_figure}

\end{figure}

We start with training the diffusion probabilistic model to generate ensembles of images (i.e. distance matrices) with proper fBm statistics. For that we synthesize $M=2\times10^{5}$ random fBm trajectories $\{x_i\}_{i=1}^N$ of length $N=64$ in $D=3$ using Davies-Harte algorithm \cite{davies1987tests} for three values of the Hurst parameter: $H=\frac{1}{3}$ (subdiffusion), $H=\frac{1}{2}$ (normal diffusion) and $H=\frac{2}{3}$ (superdiffusion). 
Then we train the DDPM for the problem of unconditional image generation on the dataset of  complete EDMs $A=\{a_{ij}\}$ (see Eq. 1), corresponding to the synthesized fBm trajectories (i.e., a separate round of training for each Hurst parameter). 
Figure 1(a) demonstrates the generated distance matrices for the three representative values of $H$ with the respective snapshots of trajectories.
Qualitatively, with decrease of $H$ trajectories turn to be more compact, enriching the distance matrices with local patterns.

To make sure that the generated ensemble of EDMs has inherited the proper fBm statistics, we compute the ensemble-averaged scaling of the mean end-to-end squared distance $\langle x^2_H(s) \rangle^{1/2}$ for segments of length $s$ by averaging along the diagonal $s$ in matrices from the generated ensemble.
Figure 1(c) demonstrates that the DDPM model learns the correct memory exponent $H$ in all the regimes within the statistical error. Some subtle deviations at small $H$ can be attributed to insufficient training. The probability densities of the rescaled pairwise distance $x/\langle x^2_H(s) \rangle^{1/2}$ collapse on the standard Gaussian independently of the Hurst parameter and segment sizes $s$, in full agreement with Eq.\ref{gauss}. Additionally, Table 1 shows that the FID score between the generated and the ground truth ensembles is quite low with the equivalent noise level not exceeding $2\%$, indicating that the diffusion generates a statistically representative ensemble of EDMs with the preset Hurst parameter. Also, the fraction of the first 5 singular values clearly demonstrates that the rank of the generated matrices does not exceed $r=5$, as it is should be for EDMs in $D=3$.
All together, the unconditional generation of distance matrices by the pre-trained DDPM reproduces the statistical properties of the ensemble of EDM images corresponding to fBm trajectories, confirming that diffusion generative models are able to learn strong algebraic correlations in the training data (see Eq. \ref{corr}). 

\begin{table}[ht]
\begin{center}
\scriptsize
\caption{DDPM unconditional generation. The FID is calculated between an ensemble of complete fBm distance matrices, which are generated by the Davies-Harte algorithm, and the unconditional DDPM samples corresponding to three different Hurst parameter values. The dimension of the InceptionV3 embedding used is 64. For the reference, we compute the FID for two ensembles of fBm distance matrices, both of which are generated independently using the Davies-Harte algorithm. The noised FID is then calculated between the exact fBm ensemble and the fBm ensemble with multiplicative noise. The level of multiplicative noise is determined to match the FID values of DDMP. The last row reflects the contribution of the first $r=5$ singular values of $A$ in the nuclear norm.}

\begin{tblr}{hlines, vlines, rows={5mm}, colspec={*{5}{Q[17mm]} Q[24mm]}}
\diagbox[width=20mm,height=5mm]{Hurst}{Metric} & \textbf{FID DDPM} & \textbf{FID reference} & \textbf{FID ref. noised} & \textbf{Equiv. noise, \%} & $\frac{\sqrt{\sum_{i=1}^{5} \lambda^{2}_{i}}}{\sqrt{\sum_{i=1}^{64} \lambda^{2}_{i}}}$ \\
\text{$H = 1/3$} & $0.097 \pm 0.012$ & $0.033 \pm 0.010$ & $0.098 \pm 0.013$ & $1.68 \pm 0.15$ & $0.998 \pm 0.001$ \\
\text{$H = 1/2$} & $0.087 \pm 0.013$ & $0.037 \pm 0.010$ & $0.083 \pm 0.011$ & $0.356 \pm 0.003$ & $0.9997 \pm 0.0003$ \\
\text{$H = 2/3$} & $0.088 \pm 0.023$ & $0.044 \pm 0.011$ & $0.094 \pm 0.015$ & $0.102 \pm 0.001$ & $0.9997 \pm 0.0003$ \\
\end{tblr}

\end{center}
\end{table}

\subsection*{Inpainting of incomplete EDMs}

As the diffusion model fairly captures and reproduces the intrinsic correlations in fBm trajectories, we next ask whether the conditional generation (inpainting) of the pre-trained DDPM is able to optimally fill missing data in incomplete EDMs with memory. For the inpainting problem, we generate symmetric binary masks with a given sparsity by sampling a Bernoulli random variable. The probability of getting a $0$ is equal to the missing ratio $\mu$ for each element in the upper triangle of $A$. Then, the partially known distance matrices $\tilde{A}$ are generated by multiplying the fully-known distance matrix $A$ by the corruption mask.

In this work, we test diffusion-based methods such as DDPM, DDNM \cite{wang2022zero}, DDRM \cite{kawar2022denoising}, and RePaint \cite{lugmayr2022repaint}. These methods only need a pre-trained diffusion model as the generative prior, but we stress that DDNM, DDRM, and RePaint additionally require knowing the corruption operators at both training and generation. In our case, the corruption operator is a known corruption mask, which helps diffusion to inpaint. Methods DDNM and RePaint use a time-travel trick (also known as resampling in \cite{lugmayr2022repaint}) for better restoration quality, aimed at intense inpainting with a huge mask, but they can be adversarial at small missing ratio $\mu$. 
It was shown in \cite{wang2022zero} that DDNM generalizes DDRM and RePaint, but in our paper, we follow the convention that DDNM is a model with parameters, where the travel length and the repeat times are both set to 3, and for RePaint, we use a number of resamplings set to 10. We further discuss the differences between these methods in the Appendix.

\begin{table}[h]
\begin{center}
\scriptsize
\caption{Comparison of different inpainting methods in EDM completion. The FID is calculated between an ensemble of distance matrices, which are generated by the Davies-Harte algorithm, and reconstructed samples corresponding to three different sparsity values $\mu$. The dimension of the InceptionV3 embedding used for FID is 64. The rank measures the contribution of the first $r=5$ singular values of the reconstructed matrix in the nuclear norm.}
\begin{tblr}{hlines,vlines, cell{2}{1}={r=3}{l}, cell{5}{1}={r=3}{l}, cell{8}{1}={r=3}{l}, rows = {5mm}, colspec = {
Q[ 8mm] Q[ 11mm] *{5}{Q[ 18mm]}
}}

Sparsity & \diagbox[width=15mm,height=6mm]{Metric}{Method} & \textbf{RePaint} & \textbf{DDRM} & \textbf{DDNM} & \textbf{DDPM} & \textbf{Database search}\\
\hline
\texttt{$\mu=0.25$} & RMSE $\downarrow$ & $0.49 \pm 0.02$ & \textbf{0.170} $\pm$ \textbf{0.017} & $0.211 \pm 0.018$ & $0.313 \pm 0.023$ & $1.12 \pm 0.12$ \\
& FID $\downarrow$ & $0.0446 \pm 0.0026$ & \textbf{0.013} $\pm$ \textbf{0.0017} & $0.027 \pm 0.0015$ & $0.0235 \pm 0.0019$ & $1.225 \pm 0.009$ \\
& Rank $\uparrow$ & $0.858 \pm 0.025$ & $0.853 \pm 0.023$ & $0.854 \pm 0.022$ & $0.849 \pm 0.025$ & $0.65 \pm 0.05$ \\
\hline

\texttt{$\mu=0.5$} & RMSE $\downarrow$ & $0.54 \pm 0.04$ & \textbf{0.241} $\pm$ \textbf{0.027} & $0.325 \pm 0.027$ & $0.55 \pm 0.05$ & $1.61 \pm 0.18$ \\
& FID $\downarrow$ & $0.053 \pm 0.003$ & \textbf{0.018} $\pm$ \textbf{0.002} & $0.053 \pm 0.002$ & $0.0246 \pm 0.0007$ & $1.79 \pm 0.01$ \\
& Rank $\uparrow$ & $0.86 \pm 0.025$ & $0.854 \pm 0.025$ & $0.853 \pm 0.025$ & $0.843 \pm 0.030$ & $0.63 \pm 0.05$ \\
\hline

\texttt{$\mu=0.75$} & RMSE $\downarrow$ & $0.68 \pm 0.06$ & \textbf{0.42} $\pm$ \textbf{0.04} & $0.56 \pm 0.04$ & $1.23 \pm 0.18$ & $1.97 \pm 0.22$ \\
& FID $\downarrow$ & $0.075 \pm 0.003$ & \textbf{0.034} $\pm$ \textbf{0.003} & $0.116 \pm 0.002$ & $0.096 \pm 0.003$ & $1.25 \pm 0.011$ \\
& Rank $\uparrow$ & $0.863 \pm 0.025$ & $0.854 \pm 0.027$ & $0.854 \pm 0.027$ & $0.82 \pm 0.04$ & $0.65 \pm 0.05$ \\
\end{tblr}
\label{tab:reconstruct_metrics}
\end{center}
\end{table}

Alongside the (i) diffusion based methods DDPM, DDNM, DDRM, RePaint for inpainting \cite{ho2020denoising}, we test other approaches (see Figure 2). (ii) FISTA is the classical low-rank completion method that exactly recovers unknown distances in the case when such recovery is unique. (iii) Trajectory optimization relies on the direct gradient optimization of 3D coordinates, incorporating a prior that the reconstruction $A$ is an EDM with rank $D+2=5$ in general configuration. (iv) The nearest neighbor (NN) method naively fills unknown matrix elements with the closest known distances in the matrix, assuming that close elements are similar. Finally, we also test the (v) database search approach, which fills unknown elements using the entries of the most similar EDM in a pre-generated database (see Methods for the details).

Visual inspection of the inpainted matrices shows remarkable results of the diffusion-based inpainting over classical EDM completion approaches at high sparsity $\mu$ (Figure 2, Figure S5). For systematic comparison of different methods, we measure RMSE and FID scores between exact distance matrices and reconstructed ones for 100 values of sparsity, equally spaced from $\mu=0.01$ to $\mu=0.99$, see Figure 3. To check for the uniqueness of the EDM completion, for each missing ratio $\mu$ we plot a fraction of uniquely recoverable EDMs that pass the rigidity test (see the previous section). We find that in the regime of small $\mu$, where the solution is unique, RMSE of the reconstruction by FISTA is the smallest, and it converges to RMSE of diffusion-based inpainting when the uniqueness is effectively lost at $\mu \approx 0.6-0.8$. However, low FID values of the inpainting similar to the ones of FISTA indicate that the diffusion approach yields correct distributions of the matrices, while the intrinsic stochasticity of DDPM is the reason of the larger RMSE. We further find that the inpainting has a significantly smaller FID than the trajectory optimization (OPT) and outperforms in RMSE at small $\mu$. Despite that OPT has the exact rank by construction, it fails to correctly reproduce the fBm-like statistics of the ensemble at small $s$ (see Figure S2). Similar behaviour of the metrics is observed for $H=1/3$ and $H=2/3$ (Figures S3,S4). Interestingly, among different diffusion-based inpainting schemes, DDRM has the smallest RMSE in a wide range of missing ratio $\mu$, as can be seen in Figure S6 and Table 2. At some point DDRM performance decreases, while DDPM and DDNM generally better behave at very high missing ratios. Our numerical experiments thus demonstrate that DDRM method has the highest precision on the fBm benchmark in the wide range of sparsity.

\subsection*{Diffusion vs Database search}

A recent study has reported that diffusion can memorize and generate examples from the training dataset \cite{carlini2023extracting,somepalli2023diffusion}. However, already for the unconditional generation and for various training database sizes we observe that the FID score between the matrices generated by diffusion and the train dataset is similar to the FID between the generated matrices and an independently synthesized set of distance matrices (Table 3). This suggests that DDPM is able to generalize rather than memorize the training examples.

\vspace{-1em}
\begin{table}[ht]
\begin{center}
\small
\caption{FID(train, generated) and FID(test, generated) for uncodnitional DDPM samples for diffusion trained on databases of different sizes. }
\begin{tblr}{hlines,vlines,rows = {4mm}, colspec = {
Q[ 22mm] Q[12mm] *{5}{Q[ 12mm]}}}

\textbf{Methods} & DB 500 & DB 1k & DB 2k  & DB 5k & DB 20k & DB 200k \\
\hline
\textbf{FID(train, gen)} & $ 0.90 \pm 0.24$& $0.56 \pm 0.04$ & $0.341 \pm 0.006$ & $0.257 \pm 0.008$ & $0.248 \pm 0.008$ & $0.0875 \pm 0.0005$ \\
\textbf{FID(test, gen)} & $ 0.91 \pm 0.23$& $0.53 \pm 0.03$& $0.335 \pm 0.008$  & $0.257 \pm 0.008$ & $0.251 \pm 0.002$ & $0.0874 \pm 0.0007$ \\

\end{tblr}
\label{tab:res_comb}
\end{center}
\end{table}

\vspace{-1em}

Next, for the problem of conditional generation we find that different schemes of the diffusion inpainting significantly outperform the database search in all sparsity regimes, see Table 2 and Figure 3. Obviously, in contrast to the inpainting completion that has low rank, the matrix completed with the database would have essentially higher rank (lower fraction of the first 5 singular values, as shown in Table 2). Furthermore, Figure 3(a)-(b) importantly shows that for various database sizes $M$, the diffusion-based inpainting displays \textit{a different convexity} of RMSE plots and different behaviour of FID. These results clearly suggest that generation by diffusion \textit{qualitatively} differs from the database search, even for the largest database size $M=10^5$, when $MN^2$ approaches the number of parameters of the diffusion model ($\approx 10^8$).

We find that the FID of the database search generation weakly depends on the database size $M$. The respective slope of FID with $M$ in the double-log scales (FID $\sim M^{-\gamma}$) seems to be independent of the missing ratio $\mu$ in the regime where the partial graph remains rigid, see Figure 3(c).
By collapsing the plots for different
$\mu$ together we find the power-law dependence of FID on $\mu$, i.e. FID $\sim \mu^a$ with $a \approx 1.41 \pm 0.06$. For the set of collapsed data at different $\mu$ we further estimate the optimal power-law exponent $\gamma \approx 0.026 \pm 0.003$ (Figure S7). 
\begin{equation}
\text{FID}_{\text{database}}(M, \mu) \sim \mu^a M^{-\gamma}, \ a \approx 1.41, \ \gamma \approx 0.026
\label{fid_scaling}
\end{equation}
Then we extrapolate the scaled FID to larger database sizes $M$ to find the effective $M^*$ needed to reproduce the correct distribution of EDMs.
Using the FID of the diffusion inpainting as a reference, we numerically obtain a range of magnitudes corresponding to the effective database size 
\begin{equation}
\log(M^*)|_{\text{exp}} = 76.7 \pm 10.3.
\label{exp}
\end{equation}

Such a giant size of the effective database $M^*$ as compared to the amount of parameters in the diffusion model is explained by the exponential dependence of $M^*$ on the trajectory length $N$. Indeed, for a single trial, the expected negative log-likelihood of a random variable $x$ from a standard normal distribution is $E[-\ln(p(x))] = \ln(\sqrt{2\pi}) + \frac{1}{2}$. For the purpose of counting we assume the trajectories are drawn on the lattice with the fixed step size and fixed origin, then, the whole ensemble of distance matrices can be generated with $\sim 2(N-1)$ Gaussian random variables. Therefore, the total entropy of the ensemble, or the log of the effective database size, reads
\begin{equation}
\log(M^*)|_{\text{theory}} = \frac{2(N-1)}{\ln(10)}\left(\ln(\sqrt{2\pi}) + \frac{1}{2}\right) \approx 78.89
\label{th}
\end{equation}
As a caveat, note that this theoretical result provides the upper bound for $M^*$, and it should be reduced upon taking into account the distance-preserving transformations. Nevertheless, comparing Eq. \ref{exp} with Eq. \ref{th} we find a remarkable agreement.

\section*{Filling missing data in the FISH dataset using the fBm benchmark}

\begin{figure}[H]
\centering    
\includegraphics[width=\textwidth]{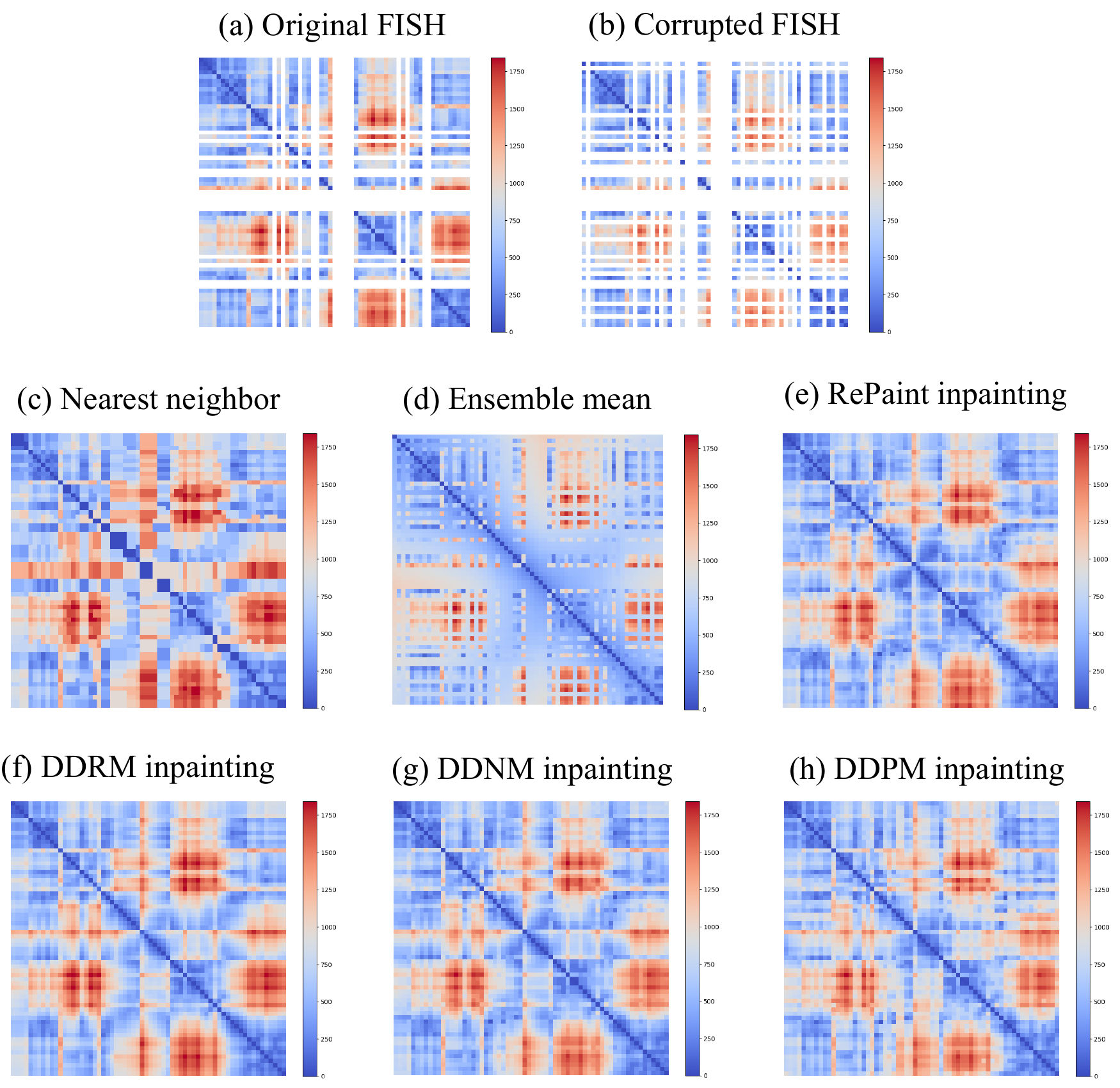}    
\caption{Inpainting of chromosome distance matrices from a FISH experiment. (a) Original experimental matrix with 15 missing rows and columns. (b) Corrupted experimental matrix with 10 rows and columns additionally masked. The masking is needed in order to evaluate RMSE of various inpainting methods at the known (masked) values. The resulting sparsity is $\mu=0.63$. (c)-(h) Inpainting methods, as indicated. For the ensemble mean (d) the average value is taken over 670 single cell distance matrices, where the corresponding matrix element is known. Diffusion-based methods (e)-(h) exploit the pre-trained diffusion model with the Hurst parameter $H=1/3$. The colorbars show the range of pairwise distances between chromosomal loci in nm. The data is shown for cell 343, see Tables S2-S3 in the Supplementary Information.}
\label{fig:figure_01}    
\end{figure}

As an application of the pre-trained fBm diffusion model, we briefly discuss the results of imputation of missing data in single cell matrices of pairwise distances between chromosomal segments (see Figure 4) obtained from microscopy experiments (Fluorescence In Situ Hybridization, FISH \cite{Bintu18}). The corresponding public folder with the raw data is available at the Github page of the paper (https://github.com/BogdanBintu/ChromatinImaging). The dataset represents the 3D coordinates of 30kb segments on a human chromosome 21 (the human colon cancer cell line, HCT116) measured using the multiplex microscopy. Noticeably, some of the coordinates are missed (nan values in the data). For our purposes of the restoration of missed coordinates we used the data with auxin that supposedly corresponds to the condition with no cohesin-mediated loops. Indeed, in vivo data shows that without loops chromosomes exhibit \textit{fractal} statistics, thus justifying the use of fBm trajectories in the training phase of our benchmark \cite{Polovnikov18,Polovnikov23}. We take the 2Mb-long segment from 28Mb to 30Mb for the analysis, the corresponding file name is "HCT116\_chr21-28-30Mb\_6h auxin.txt".

First, using the raw data we reproduce the fractal scaling of chromosomal folding \cite{Lieberman09,Polovnikov23,Polovnikov18}, i.e. $\langle x^2(s) \rangle^{1/2} \sim s^{1/3}$, see Figure S8. We stress that such a behaviour corresponding to the fractal dimension $d_f=3$ is a result of depleted cohesin from the cell nucleus, which is known to extrude loops on chromosomes (see our previous works \cite{Polovnikov23,Polovnikov23pre,Polovnikov23jetp} for the biophysical model showing how cohesin loops break the scale-invariance). Thus, to inpaint the missing data we decided to use the diffusion model trained on the fBm ensemble with $H=1/3$ (the fractal dimension is the inverse of the Hurst parameter). Figure 4 shows the resulting matrices obtained using various inpainting methods run on a particular FISH dataset (cell 343, see Tables S2-S3 for the data).


\begin{table}[h]
\begin{center}
\small
\caption{Reconstruction of chromatin (FISH) distance matrices. Average metrics over 670 single cells are shown}
\begin{tblr}{hlines,vlines,rows = {4mm}, colspec = {
Q[ 15mm] Q[18mm] *{5}{Q[ 15mm]}}}

\textbf{Methods} & Ens. mean & NN  & DDPM & RePaint & DDNM & DDRM \\
\hline
\textbf{RMSE, nm} & $147.4 \pm 5.2$ & $111.4 \pm 3.2$ & $97.2 \pm 2.5$ & $98.3 \pm 2.7$ & $85.1 \pm 2.8$ & \textbf{84.2} $\pm$ \textbf{2.8} \\
\textbf{Rank} & $0.752 \pm 0.025$& $0.79 \pm 0.03$ &  $0.79 \pm 0.04$ & $0.82 \pm 0.04$ & $0.82 \pm 0.03$ & $0.82 \pm 0.03$ \\

\end{tblr}
\label{tab:res_comb}
\end{center}
\end{table}

To measure the performance of the DDPM inpainting in comparison to other methods we chose 670 cells (out of 7380 cells) in the dataset that have exactly 15 missing rows and columns. This corresponds to sparsity $\mu'=0.29$. In order to compute RMSE (Table 4) we additionally dropped 10 rows and columns from the matrices resulting in $\mu=0.63$. We then imputed the missing distances using the standard bioinformatics approaches (nearest neighbor, ensemble mean) and various diffusion-based inpainting methods (DDRM, DDNM, DDPM, RePaint). The nearest neighbor approach relies on filling the unknown distances with the nearest neighbour in the same matrix (cell). The ensemble mean approach fills the missing entry with the corresponding average over the cells where this element is known. The average RMSE was computed for each imputed cell over the known values in the dropped 10 columns and rows. Note that since the \textit{entire} rows and columns are missing in FISH distance matrices, precise EDM completion algorithms such as FISTA or trajectory optimization (OPT) are not applicable here. 

Consistently with numerical experiments, the Table 4 demonstrates that the DDRM inpainting trained on the fBm benchmark is superior over other diffusion-based and bioinformatics approaches. It should be noted that it has a comparable RMSE and rank with DDNM inpainting, while RePaint and DDPM behave slightly worse (the resulting RMSE is more than 10\% larger). This is to be compared with other approaches, such as filling the missing distances using the nearest neighbor pixel from the same matrix (NN) or using the average over the cells where this matrix element is known (Ens. mean). These clearly naive approaches behave significantly worse both in the rank and RMSE. This observation highlights that the fBm benchmark for diffusion-based inpainting shows evidently better performance on a biological dataset than canonical bioinformatics approaches.

\section*{Conclusion}

By treating the Euclidean distance matrices as images, in this paper we demonstrated that diffusion probabilistic model 
can learn the essential large-scale correlations in the distance matrices of the ensemble of fBm trajectories for various memory exponents $H$. Based on this observation, 
we apply the diffusion-based inpainting trained on the fBm benchmark for the problem of EDM completion, exploring diffusion inpainting methods at various sparsity parameters. Using our benchmark we observe that the diffusion behaves drastically different from the database search with the database size similar to the number of parameters of the diffusion model. We provide a theoretical argument for the effective database size explaining such a qualitative difference and verify it in numerical experiments.
We further show that the diffusion-based inpainting not only learns the latent representation of the distance matrices, but also manages to properly reproduce the statistical features of the fBm ensemble (the memory exponent). Application of the pre-trained fBm benchmark for the microscopy-derived dataset of pairwise spatial distances between chromosomal segments demonstrates its superiority in reconstructing the missing distances over the standard approaches widely used in bioinformatics. We thus expect that other chromosomal datasets obtained in high-throughput experiments (such as Hi-C) that can be represented as matrices would benefit from the proposed fBm benchmark.

It should be noted that the instance-based metrics of the inpainting quality, such as RMSE, PSNR and SSIM, are well-grounded only when the inpainting ground truth exists (for example, for the EDM dataset at low $\mu$). In the image domain, however, masking even a relatively small area can lead to several equally possible inpainting results. For example, if the facial features (such as eyes) are masked, the output could include natural variations of the face (e.g., different eye colors).  Thus, for conventional image datasets, such as CIFAR-10, ImageNet, or LAION, the inpainting problem is rather ill-posed. In contrast to the natural images, the EDM dataset we proposed in this paper uniquely allows to study the performance of the generative models at the instance level. Additionally, this enables one to make systematic comparisons between the diffusion inpainting and database search results, revealing the fundamental generalization capabilities of the diffusion models.

As the main limitation of our fBm benchmark, we stress out that it is not directly applicable to noisy EDMs, i.e. we assume that the known pairwise distances are exact. This should be seriously taken into account upon application of the benchmark to real datasets. Also the wild type chromosomes (\textit{with} cohesin present in the cell) show a distinctively non-fractal statistics, thus requiring an adequate modification of the benchmark for the purposes of the data imputation. Other caveats are discussed in the flow of the paper.

\section*{Methods}

We conducted all experiments on a workstation equipped with two NVIDIA RTX 4090 graphics cards and 256 GB of RAM. The DDPM model was parameterized using the 
UNet2DModel from the Diffusers library, and we trained the model for 100 epochs on datasets of generated fBm trajectories for dataset sizes ranging from 500 to 200,000 distance matrices (500, 1000, 2000, 5000, 20000 and 200000), see Table 3. For other experiments we used the diffusion model trained on the dataset of 200k distance matrices. For the inpaitning experiments we used 200 generation steps. The source code for the experiments is available at \url{https://github.com/alobashev/diffusion_fbm}.

\subsection*{DDPM inpainting}

Denoising Diffusion Probabilistic Models (DDPM) generate data by reversing a forward noising process \cite{ho2020denoising,song2020score}. The forward process incrementally adds Gaussian noise to an initial sample $x_0 \sim p_{\text{data}}$ across a sequence of steps, following a variance schedule ${\beta_1, ..., \beta_T}$. Let's denote $\alpha_t := 1- \beta_t$ and cumulative $\bar{\alpha}_t:= \prod_{i = 1}^t\alpha_i$. When the schedule is properly set and $T$ is sufficiently large, the noised sample becomes indistinguishable from the pure noise $\mathcal{N}(0, I)$. The forward process is characterized by the distribution:

$$ q(x_t | x_{t - 1}) := \mathcal{N}(x_t; \sqrt{1 - \beta_t}x_{t - 1}, \beta_tI), \ \ \ \ \ \ \ q(x_{1:T}|x_0) = \prod_{t = 1}^T q(x_t | x_{t - 1})$$

Conversely, the backward process aims to denoise a given sample, gradually recovering the original object from the noise. The diffusion model, a generative model with latent variables, is defined as:

$$p_\theta(x_0) := \int p_\theta(x_{0:T}) dx_{1:T}$$

The joint distribution $p_\theta(x_{0:T})$ forms a reverse Markov chain:

$$p(x_{0:T}) = p(x_T) \prod_{t = 1}^Tp_{\theta}(x_{t-1}|x_t) \ \ \ \ \ \ \ \ \ p(x_{T})=\mathcal{N}(x_T; 0, I)$$
$$ p_{\theta}(x_{t - 1}|x_t):= \mathcal{N}(x_{t - 1}; \mu_{\theta}(x_t, t), \Sigma_{\theta}(x_t, t))$$

The variational lower bound (VLB) is optimized during training, which involves the distribution $q(\mathbf{x}_{t-1} | \mathbf{x}_t, \mathbf{x}_0)$:

\begin{equation}  
\begin{split}  
L_{\text{VLB}} = \mathbb{E}_q [\underbrace{D_\text{KL}(q(x_T | x_0) \parallel p(x_T))}_{L_T} + \\  
+ \sum_{t=2}^T \underbrace{D_\text{KL}(q(x_{t-1} | x_t, x_0) \parallel p_\theta(x_{t-1} | x_t))}_{L_{t-1}}] - \underbrace{\log p_\theta(x_0 | x_1)}_{L_0}  
\end{split}  
\end{equation}  

The VLB loss can be simplified to another loss, which we use for the training:

$$L^{\text{simple}} = \mathbb{E}_{x_0, \epsilon, t}\bigg[ |\epsilon - \epsilon_{\theta}(x_t, t)|^2\bigg]$$

To sample from the model (backward process), we use $\mu_{\theta}(x_t, x_0)$, which is derived from $\epsilon_{\theta}(x_t, t)$ as:
$$
\mu_{\theta}(x_{t}, t) = \frac{1}{\sqrt{\alpha_{t}}} \bigg[ x_{t} - \frac{\beta_{t}}{\sqrt{1-\bar{\alpha}_{t}}} \epsilon_{\theta}(x_t, t) \bigg]
$$

In practice, we aim for faster sampling with a number of steps that is lower than that is required for the training. If the Markov chain is shortened and we use only a subsequence $S_{i}$ of the diffusion steps, the original schedule of variances is adjusted to maintain the same marginal distributions $q^{\text{new}}(x_i) = q(x_{S_i})$. From that condition, we can derive the betas for the new (shorter) diffusion process as follows:

\begin{align}
    \beta_i^{\text{new}} = 1 - \frac{
        \bar{\alpha}_{S_i}
    }{
        \bar{\alpha}_{S_{i-1}}
    }
\end{align}

The sampling process remains unchanged, only requiring a modification of the pretrained $\epsilon$ inputs to $\epsilon(x_i, S_i)$, where $S_i$ is the corresponding timestep in the original chain. We set the number of sampling iterations to $150$ for all our experiments.

The DDPM have recently been adapted into a variants, one of them known as Repaint \cite{lugmayr2022repaint}, which enhances the inpainting process by iteratively refining the generated samples. Unlike the original DDPM approach, which progresses through a single pass of forward and backward steps, Repaint employs a strategy where the reverse diffusion process is performed multiple times, or in "loops," to achieve higher fidelity results. 
Given the mask $m$, DDPM enables us to solve the inpainting problem as follows:
\begin{align}
x_{t-1}^{\text{known}} &\sim \mathcal{N}(\sqrt{\alpha_t}x_0, (1 - \bar{\alpha}t)I)  \\
x_{t-1}^{\text{unknown}} &\sim \mathcal{N}(\mu_{\theta}(x_t, t), \Sigma_{\theta}(x_t, t)) \\
x_{t-1} &= m \odot x_{t-1}^{\text{known}} + (1 - m) \odot x_{t-1}^{\text{unknown}}
\end{align}

The known pixels are sampled from the initial image $m \odot x_0$, while the unknown pixels are sampled from the model conditioned on the previous iteration $x_t$. The samples are then merged to form the new image $x_{t-1}$ using the mask. This is known as the RePaint procedure, which we use for the distance matrix inpainting. It is important to note that DDPM is equivalent to Repaint when the length of the time-loops is set to one.

Given a trained DDPM, one can perform image inpainting by applying a quadratic potential to the known distance matrix values $\hat{X}_{t}$, which are noised at the same level as $X_{t}$. In the reverse diffusion process, the gradient of this potential is added to the score function. This procedure is reduced to a projection applied after every sampling iteration:
\begin{equation}
X_{t} = X_{t} + B \odot (\hat{X}_{t} - X_{t})
\end{equation}
Here, $B$ represents the corruption mask.

\begin{algorithm}[H]
\caption{Reverse Diffusion Process of DDPM Unconditional Generation}
\label{alg:ddpm}
\textbf{Require}: None
\begin{algorithmic}[1]
\State $\mathbf{x}_{T} \sim \mathcal{N}(\mathbf{0}, \mathbf{I})$
\For{$t = T$ \textbf{to} $1$}
    \State $\boldsymbol{\epsilon} \sim \mathcal{N}(\mathbf{0}, \mathbf{I})$ \textbf{if} $t > 1$, \textbf{else} $\boldsymbol{\epsilon} = \mathbf{0}$
    \State $\mathbf{x}_{t-1} = \frac{1}{\sqrt{\alpha_{t}}} \left( \mathbf{x}_{t} - \mathcal{Z}_{\boldsymbol{\theta}}(\mathbf{x}_{t}, t) \frac{\beta_{t}}{\sqrt{1-\bar{\alpha}_{t}}} \right) + \sigma_{t}\boldsymbol{\epsilon}$
\EndFor
\State \Return $\mathbf{x}_{0}$
\end{algorithmic}
\end{algorithm}

\begin{algorithm}[H]
\caption{Reverse Diffusion Process of DDPM Inpainting}
\label{alg:repaint}
\textbf{Require}: The masked image $\mathbf{y}$, the mask $\mathbf{A}$
\begin{algorithmic}[1]
\State $\mathbf{x}_{T} \sim \mathcal{N}(\mathbf{0}, \mathbf{I})$
\For{$t = T$ \textbf{to} $1$}
    \State $\boldsymbol{\epsilon}_{1}, \boldsymbol{\epsilon}_{2} \sim \mathcal{N}(\mathbf{0}, \mathbf{I})$ \textbf{if} $t > 1$, \textbf{else} $\boldsymbol{\epsilon}_{1}, \boldsymbol{\epsilon}_{2} = \mathbf{0}$
    \State $\mathbf{y}_{t-1} = \sqrt{\bar{\alpha}_{t-1}}\mathbf{y} + \sqrt{1-\bar{\alpha}_{t-1}}\boldsymbol{\epsilon}_{1}$ \label{alg:repaint_yt}
    \State $\mathbf{x}_{t-1} = \frac{1}{\sqrt{\alpha_{t}}} \left( \mathbf{x}_{t} - \mathcal{Z}_{\boldsymbol{\theta}}(\mathbf{x}_{t}, t)\frac{\beta_{t}}{\sqrt{1-\bar{\alpha}_{t}}} \right) + \sigma_{t}\boldsymbol{\epsilon}_{2}$
    \State $\mathbf{x}_{t-1} = \mathbf{y}_{t-1} + (\mathbf{I} - \mathbf{A})\mathbf{x}_{t-1}$ \label{alg:repaint_rnd}
\EndFor
\State \Return $\mathbf{x}_{0}$
\end{algorithmic}
\end{algorithm}

\subsection*{RePaint}
RePaint \cite{lugmayr2022repaint} solves noise-free image inpainting problems, where $\mathbf{n}=0$ and $\mathbf{A}$ represents the mask operation. RePaint first create a noised version of the masked image $\mathbf{y}$
\begin{equation}
    \mathbf{y}_{t-1} =\mathbf{A}( \sqrt{\bar{\alpha}_{t-1}}\mathbf{y}+\sqrt{1-\bar{\alpha}_{t-1}}\boldsymbol{\epsilon}), \quad \boldsymbol{\epsilon}\sim \mathcal{N}(0,\mathbf{I}).
    \label{eq:repaint yt-1}
\end{equation}
Then uses $\mathbf{y}_{t-1}$ to fill in the unmasked regions in $\mathbf{x}_{t-1}$:
\begin{equation}
    \mathbf{x}_{t-1} = \mathbf{y}_{t-1} + (\mathbf{I} -\mathbf{A})\mathbf{x}_{t-1},
    \label{eq:repaint xt-1}
\end{equation}
Besides, RePaint applies an ``back and forward" strategy to refine the results. Algo.~\ref{alg:repaint} shows the algorithm of RePaint.

\begin{algorithm}[H]
\caption{Reverse Diffusion Process of RePaint}
\label{alg:repaint}
\textbf{Require}: The masked image $\mathbf{y}$, the mask $\mathbf{A}$
\begin{algorithmic}[1]
\State $\mathbf{x}_{T} \sim \mathcal{N}(\mathbf{0}, \mathbf{I})$
\For{$t = T$ \textbf{to} $1$}
    \For{$s = 1$ \textbf{to} $S_{t}$}
        \State $\boldsymbol{\epsilon}_{1}, \boldsymbol{\epsilon}_{2} \sim \mathcal{N}(\mathbf{0}, \mathbf{I})$ \textbf{if} $t > 1$, \textbf{else} $\boldsymbol{\epsilon}_{1}, \boldsymbol{\epsilon}_{2} = \mathbf{0}$
        \State $\mathbf{y}_{t-1} = \sqrt{\bar{\alpha}_{t-1}}\mathbf{y} + \sqrt{1-\bar{\alpha}_{t-1}}\boldsymbol{\epsilon}_{1}$ \label{alg:repaint_yt}
        \State $\mathbf{x}_{t-1} = \frac{1}{\sqrt{\alpha_{t}}} \left( \mathbf{x}_{t} - \mathcal{Z}_{\boldsymbol{\theta}}(\mathbf{x}_{t}, t)\frac{\beta_{t}}{\sqrt{1-\bar{\alpha}_{t}}} \right) + \sigma_{t}\boldsymbol{\epsilon}_{2}$
        \State $\mathbf{x}_{t-1} = \mathbf{y}_{t-1} + (\mathbf{I} - \mathbf{A})\mathbf{x}_{t-1}$ \label{alg:repaint_rnd}
        \If{$t \neq 0$ \textbf{and} $s \neq S_{t}$}
            \State $\mathbf{x}_{t} = \sqrt{1-\beta_{t}}\mathbf{x}_{t-1} + \sqrt{\beta_{t}}\boldsymbol{\epsilon}_{2}$ \label{alg:repaint_loop}
        \EndIf
    \EndFor
\EndFor
\State \Return $\mathbf{x}_{0}$
\end{algorithmic}
\end{algorithm}

\subsection*{DDRM}
The forward diffusion process defined by DDRM \cite{kawar2022denoising} is
\begin{equation}
    \mathbf{x}_{t} = \mathbf{x}_{0} + \sigma_{t}\boldsymbol{\epsilon},\quad \boldsymbol{\epsilon}\sim\mathcal{N}(\mathbf{0},\mathbf{I})
    \label{eq:ddrm1}
\end{equation}
The original reverse diffusion process of DDRM is based on DDIM, which is
\begin{equation}
    \mathbf{x}_{t-1} = \mathbf{x}_{0} + \sqrt{1-\eta^{2}}\sigma_{t-1}\frac{\mathbf{x}_{t}-\mathbf{x}_{0}}{\sigma_{t}} + \eta\sigma_{t-1}\boldsymbol{\epsilon}
    \label{eq:ddrm2}
\end{equation}

\begin{algorithm}[H]
\caption{Reverse Diffusion Process of DDRM}
\label{alg:ddrm}
\textbf{Require}: The degraded image $\mathbf{y}$ with noise level $\sigma_{\mathbf{y}}$, the operator $\mathbf{A}=\mathbf{U}\Sigma\mathbf{V}^{\top}$, where $\mathbf{A} \in \mathbb{R}^{d\times D}$
\begin{algorithmic}[1]
\State $\mathbf{x}_{T} \sim \mathcal{N}(\mathbf{0}, \mathbf{I})$
\State $\bar{\mathbf{y}} = \Sigma^{\dagger}\mathbf{U}^{\top}\mathbf{y}$
\For{$t = T$ \textbf{to} $1$}
    \State $\boldsymbol{\epsilon} \sim \mathcal{N}(\mathbf{0}, \mathbf{I})$ \textbf{if} $t > 1$ \textbf{else} $\boldsymbol{\epsilon} = \mathbf{0}$
    \State $\bar{\mathbf{x}}_{0|t} = \mathbf{V}^{\top} \frac{1}{\sqrt{\bar{\alpha}_{t}}} \left( \mathbf{x}_{t} - \mathcal{Z}_{\boldsymbol{\theta}}(\mathbf{x}_{t}, t)\sqrt{1-\bar{\alpha}_{t}} \right)$
    \For{$i = 1$ \textbf{to} $D$}
        \If{$s_{i} = 0$}
            \State $\bar{\mathbf{x}}_{t-1}^{(i)} = \bar{\mathbf{x}}_{0|t}^{(i)} + \sqrt{1-\eta^{2}}\sigma_{t-1}\frac{\bar{\mathbf{x}}^{(i)}_{t} - \bar{\mathbf{x}}_{0|t}^{(i)}}{\sigma_{t}} + \eta\sigma_{t-1}\boldsymbol{\epsilon}^{(i)}$
        \ElsIf{$\sigma_{t-1} < \frac{\sigma_{\mathbf{y}}}{s_{i}}$}
            \State $\bar{\mathbf{x}}_{t-1}^{(i)} = \bar{\mathbf{x}}_{0|t}^{(i)} + \sqrt{1-\eta^{2}}\sigma_{t-1}\frac{\bar{\mathbf{y}}^{(i)} - \bar{\mathbf{x}}_{0|t}^{(i)}}{\sigma_{\mathbf{y}}/s_{i}} + \eta\sigma_{t-1}\boldsymbol{\epsilon}^{(i)}$
        \ElsIf{$\sigma_{t-1} \geq \frac{\sigma_{\mathbf{y}}}{s_{i}}$}
            \State $\bar{\mathbf{x}}_{t-1}^{(i)} = \bar{\mathbf{y}}^{(i)} + \sqrt{\sigma_{t-1}^{2} - \frac{\sigma_{\mathbf{y}}^{2}}{s_{i}^{2}}}\boldsymbol{\epsilon}^{(i)}$
        \EndIf
    \EndFor
    \State $\mathbf{x}_{t-1} = \mathbf{V}\bar{\mathbf{x}}_{t-1}$
\EndFor
\State \Return $\mathbf{x}_0$
\end{algorithmic}
\end{algorithm}

\subsection*{DDNM}

The forward and backward process in DDNM \cite{wang2022zero} is similar to the DDRM. 

\begin{algorithm}[H]
\caption{Reverse Diffusion Process of DDNM}
\label{alg:ddnm_appendix_version}
\textbf{Require}: The degraded image $\mathbf{y}$, the degradation operator $\mathbf{A}$ (mask), and its pseudo-inverse $\mathbf{A}^{\dagger}$.
\begin{algorithmic}[1]
\State $\mathbf{x}_{T} \sim \mathcal{N}(\mathbf{0}, \mathbf{I})$
\For{$t = T$ \textbf{to} $1$}
    \State $\boldsymbol{\epsilon} \sim \mathcal{N}(\mathbf{0}, \mathbf{I})$ \textbf{if} $t > 1$ \textbf{else} $\boldsymbol{\epsilon} = \mathbf{0}$
    \State $\mathbf{x}_{0|t} = \frac{1}{\sqrt{\bar{\alpha}_{t}}} \left( \mathbf{x}_{t} - \mathcal{Z}_{\boldsymbol{\theta}}(\mathbf{x}_{t}, t)\sqrt{1-\bar{\alpha}_{t}} \right)$
    \State $\mathbf{\hat{x}}_{0|t} = \mathbf{x}_{0|t} - \mathbf{A}^{\dagger}(\mathbf{A}\mathbf{x}_{0|t} - \mathbf{y})$
    \State $\mathbf{x}_{t-1} = \frac{\sqrt{\bar{\alpha}_{t-1}}\beta_{t}}{1-\bar{\alpha}_{t}} \mathbf{\hat{x}}_{0|t} + \frac{\sqrt{\alpha_{t}}(1-\bar{\alpha}_{t-1})}{1-\bar{\alpha}_{t}} \mathbf{x}_{t} + \sigma_{t}\boldsymbol{\epsilon}$
\EndFor
\State \Return $\mathbf{x}_{0}$
\end{algorithmic}
\end{algorithm}

\subsection*{Database Search for Inpainting}

Existence of the ground truth of the reconstruction and the respective threshold, below which the solution is unique, allows us to directly compare the reconstruction error of a generative diffusion model with the other methods. In particular, in what follows, we will test the performance of the database search, i.e. where the missing distances are extracted from the most similar matrix from the database evaluated at known distances.

It has been demonstrated that modern text-to-image diffusion models, such as StableDiffusion \cite{rombach2022high}, partially memorize samples from the training dataset \cite{carlini2023extracting,somepalli2023diffusion}. This leads to the question of whether diffusion models could function as an approximate database search.

In the context of inpainting, the database search approach leverages a pre-existing database of  distance matrices of ensemble of trajectories. The error $\varepsilon_i$, characterizing the discrepancies between the corrupted matrix and each matrix $i$ in the database, is computed as
\begin{equation}
\varepsilon_{i} = {||B \odot (A^{(DB)}_{i} - \tilde{A})||}_{F}^2,
\label{db}
\end{equation}
where $\tilde{A}$ is the partially known distance matrix, $B$ is the binary mask of known distances, and $A^{(DB)}_{i}$ is an ensemble of complete EDMs of real trajectories from the database.

Let us denote the index of the matrix from the database that provides the minimum to the error Eq. \ref{db} by $i^{*} = \text{argmin}(\varepsilon_{i})$. Then the reconstruction $\hat{A}$ is obtained by integrating information from the original corrupted matrix and the closest match from the database and is given by
\begin{equation}
\hat{A} = B \odot \tilde{A} + (1 - B) \odot A^{(DB)}_{i^{*}}.
\end{equation}
Note, that if $\tilde{A}$ belongs to the database $A^{(DB)}$ then this procedure would would give a reconstruction with zero error (if the EDM completion is unique). In other words, this inpainting procedure ideally over-fits the training data.

\subsection*{FISTA for low-rank distance matrix completion} 

Another approach to fill the unknown distances is to apply convex relaxation, by combining the minimization objective $||B \odot (A - \tilde{A})||_F^2$ with the L1 norm on $X$'s nuclear norm. This makes matrix completion a regularized least square problem

\begin{equation}  
\min_{A} ||B \odot (A - \tilde{A})||_F^2 + \beta ||A||_*
\label{eq:fista_loss}
\end{equation}  
where $\tilde{A}$ is a partially known distance matrix, $B$ is a binary mask which represents known distance matrix entities, and $A$ is the matrix to be reconstructed, $\beta$ is the regularization coefficient, and $||A||_*$ is the nuclear norm of the matrix $A$, equivalent to the sum of all $A$'s eigenvalues. 

There were proposed several methods to solve this problem, one of them is the Fast Iterative Shrinkage-Thresholding Algorithm (FISTA) \cite{beck2009fast}. FISTA operates by iteratively updating the solution via a proximal gradient method and employs Nesterov's acceleration to enhance the rate of convergence. 
FISTA serves as ground truth method for the solution of the low rank matrix completion problem in cases where solution is unique, i.e. when the partial graph defined by the mask $B$ is rigid. 

Define the singular value soft-thresholding operator as:  
\begin{equation}  
D_{\beta}(A) = U (\Sigma - \beta I)_{+} V^T  
\end{equation}  
where $A=U \Sigma V^T$ is the singular value decomposition of $A$ and $(x)_{+} := \max(x,0)$. Then the FISTA update rule is given by:  
\begin{equation}  
t_{k+1} = \frac{1+\sqrt{1+4 t_{k}^{2}}}{2}
\end{equation} 
\begin{equation}  
Z^{k+1} = A^{k+1}+\frac{t_{k}-1}{t_{k+1}}(A^{k+1}-A^{k})
\end{equation} 
\begin{equation}  
A^{k+1} = D_{\beta}( \tilde{A} \odot B + (1-B) \odot Z^{k+1} )  
\end{equation}  
and the initial approximation $A^{0}$ is the matrix $\tilde{A}$ with unknown elements filled with zeros. The stopping criterion for the iterative process is based on the ratio of the loss from Eq. \ref{eq:fista_loss} over two consecutive iterations. 

\subsection*{Inpainting using trajectory optimization}

The EDM completion problem can be tackled from the trajectory optimization perspective. The objective function to be minimized is the mean squared error (MSE) between the original and reconstructed distance matrices, computed over the known elements of the corrupted distance matrix. The loss function $L$ over the trajectory $x=\{x\}_{i=1,...,N}$ can be defined as:   
\begin{equation}   
L(x) = ||B\odot(A(x)-\tilde{A})||_{F}^2 = 
\end{equation}
\begin{equation}   
 = \sum_{||x_i-x_j||\ \text{known}}(a_{ij}(x) - \tilde{a}_{ij})^2.
\end{equation}
  
The optimization process is carried out in $n$ steps. At each step, a reconstructed distance matrix $A$ is obtained by calculating the pairwise Euclidean distance between the points in the current trajectory.  The trajectory $x$ is updated iteratively via the Adam optimizer to minimize the loss function $L$. The final output is the reconstructed distance matrix
\begin{equation}   
\hat{A} = A(x^{*}), \ \ x^{*} = \text{arg}\min L(x),
\end{equation}
which is expected to approximate the original distance matrix $\tilde{A}$ under the given constraints.

\subsection*{A greedy algorithm for the graph rigidity}

To better understand the idea of the algorithm, see Figure S1. At the first step we identify the largest clique (red nodes). For the clique we can reconstruct the full set of coordinates, as all cliques are universally rigid. At the next step we are looking for a vertex outside of the clique that has the maximum number of links to the clique, but not less than 4 (the blue node in Figure S1). If there is no such a vertex, the algorithm terminates and we conclude that the graph is not rigid. If it exists, we can determine the 3D coordinates of this vertex given the known coordinates of the vertices in the clique. In this case, we effectively know all the pairwise distances between the new blue vertex and all the red vertices in the clique, i.e. the number of vertices in the clique effectively increases by 1. We continue adding new vertices to the growing clique (green, orange) by repeating these steps, until all graph vertices join the clique. If the algorithm terminates before that, the graph is said to be not rigid.

In the following listing we provide the pseudo-code of the algorithm checking for the rigidity of any given graph. If the graph passes this check, it is rigid and allows for the unique immersion in the 3D metric space up to the distance-preserving transformations.

\begin{algorithm}[H]
\caption{Rigidity test}
\label{alg:chordality_algorithm} 
\begin{algorithmic}[1]
\Require $M$ \Comment{$M$ is the binary mask representing known elements with 0 and unknowns with 1}
\State $N \gets \text{length}(M)$ \Comment{$N$ is the number of vertices in the graph}  
\State $S_{max} \gets 0$ \Comment{Initializes the size of the largest clique found to 0}  
\State $B \gets \emptyset$ \Comment{$B$ is the best set of vertices found so far, initially empty}

\For{$i \gets 1$ to $N$}
    \State $C \gets \{i\}$
    \For{$j \gets 1$ to $N$}
        \If{$i \neq j$ \textbf{and} $M[i, j] = 0$ \textbf{and} $\forall k \in C, M[k, j] = 0$}
            \State $C \gets C \cup \{j\}$
        \EndIf
    \EndFor
    \If{\text{length}($C) > S_{max}$}
        \State $S_{max} \gets \text{length}(C)$
        \State $B \gets C$
    \EndIf
\EndFor

\While{\text{length}($B) < N$}
    \State $K_{max} \gets -1$
    \State $P_{best} \gets \text{null}$
    \For{$i \gets 1$ to $N$}
        \If{$i \notin B$}
            \State $K_{count} \gets \sum_{j \in B} [M[i, j] = 0]$
            \If{$K_{count} > K_{max}$}
                \State $K_{max} \gets K_{count}$
                \State $P_{best} \gets i$
            \EndIf
        \EndIf
    \EndFor
    \If{$K_{max} < 4$}
        \State \Return \textbf{False}
    \Else
        \State $B \gets B \cup \{P_{best}\}$
        \State $M[P_{best}, :] \gets 0$
        \State $M[:, P_{best}] \gets 0$
    \EndIf
\EndWhile
\State \Return \textbf{True}
\end{algorithmic}
\end{algorithm}

\section*{Data availability}
The source code for the experiments is available at \url{https://github.com/alobashev/diffusion_fbm}.

\bibliography{main}

\section*{Acknowledgements}

The authors are grateful to members of the laboratories of Ralf Metzler and Leonid Mirny for illuminating discussions. K.P. acknowledges the hospitality of Institute Curie, LPTMS
laboratory (Paris-Saclay University), where part of this work was done.

\section*{Author contributions statement}

K.P. conceived and conceptualized the study, A.L. and D.G. conducted the experiments, A.L. and K.P. analysed the data. All authors participated in writing the manuscript. 

\section*{Additional information}
The authors declare no competing interests.

\newpage

\section*{Supplementary Information}

\begin{figure}[h]
\centering    
\includegraphics[width=0.8\textwidth]{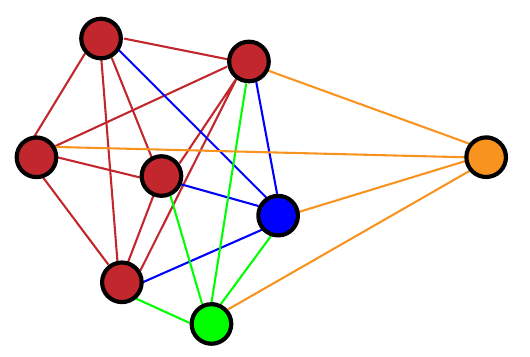}    
\caption{A sketch of the algorithm checking for rigidity. The nodes that are added at the first step are shown in red, the blue node is added at the second, the green is added at the third, the orange one is added at the fourth step.}    
\label{fig:figure_01}    
\end{figure}

\begin{figure}[H]
\centering    
\includegraphics[width=0.8\textwidth]{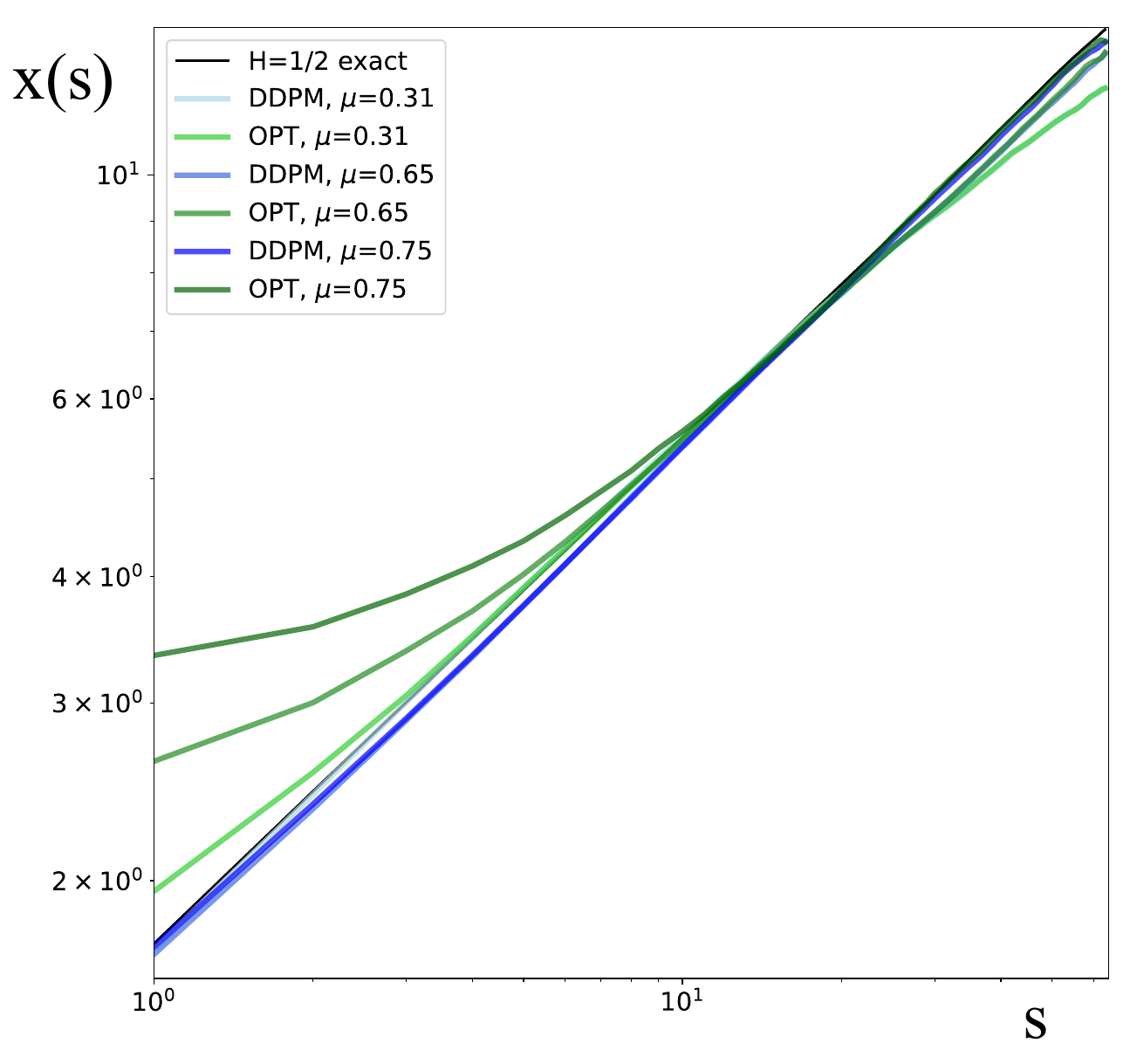}    
\caption{Scalings of the typical spatial size $x(s)$ of the trajectory segment of the contour length $s$ for the imputed EDM matrices using optimization (OPT) and DDPM inpainting (H=1/2). While the inpainting reproduces the theoretical behaviour at all scales $s$ and all $\mu$, the trajectory optimization approach tends to violate the correct scaling upon the increase of sparsity $\mu$ at small and large scales.}    
\label{fig:figure_04}    
\end{figure}

\begin{figure}[H]
\centering    
\includegraphics[width=\textwidth]{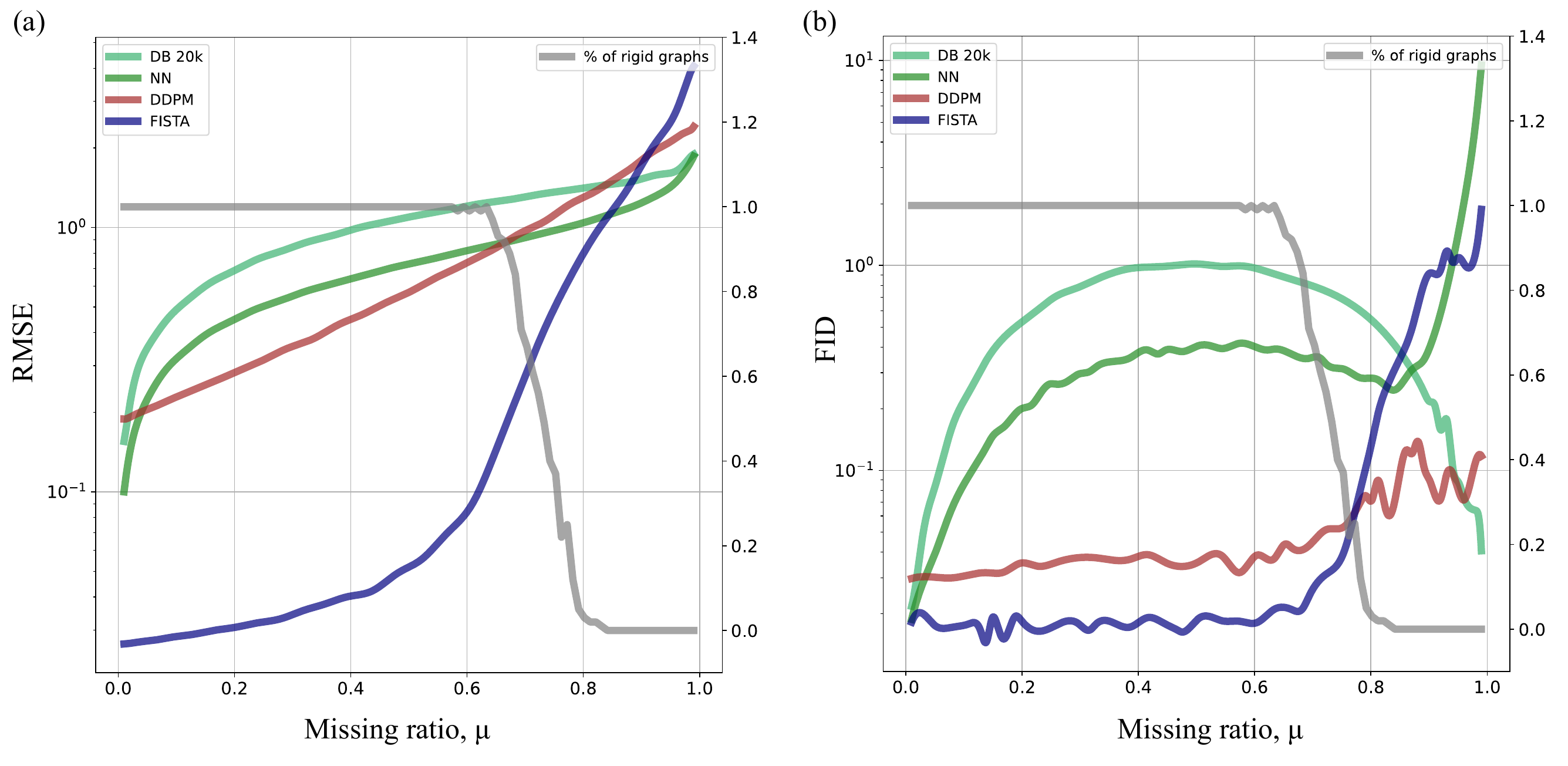}    
\caption{RMSE (a) and FID (b) plots as a function of missing ratio $\mu$ for different data imputation methods ($H=1/3$). The fraction of rigid graphs is shown in the second axis (grey).}    
\label{fig:figure_01}    
\end{figure}

\begin{figure}[H]
\centering    
\includegraphics[width=\textwidth]{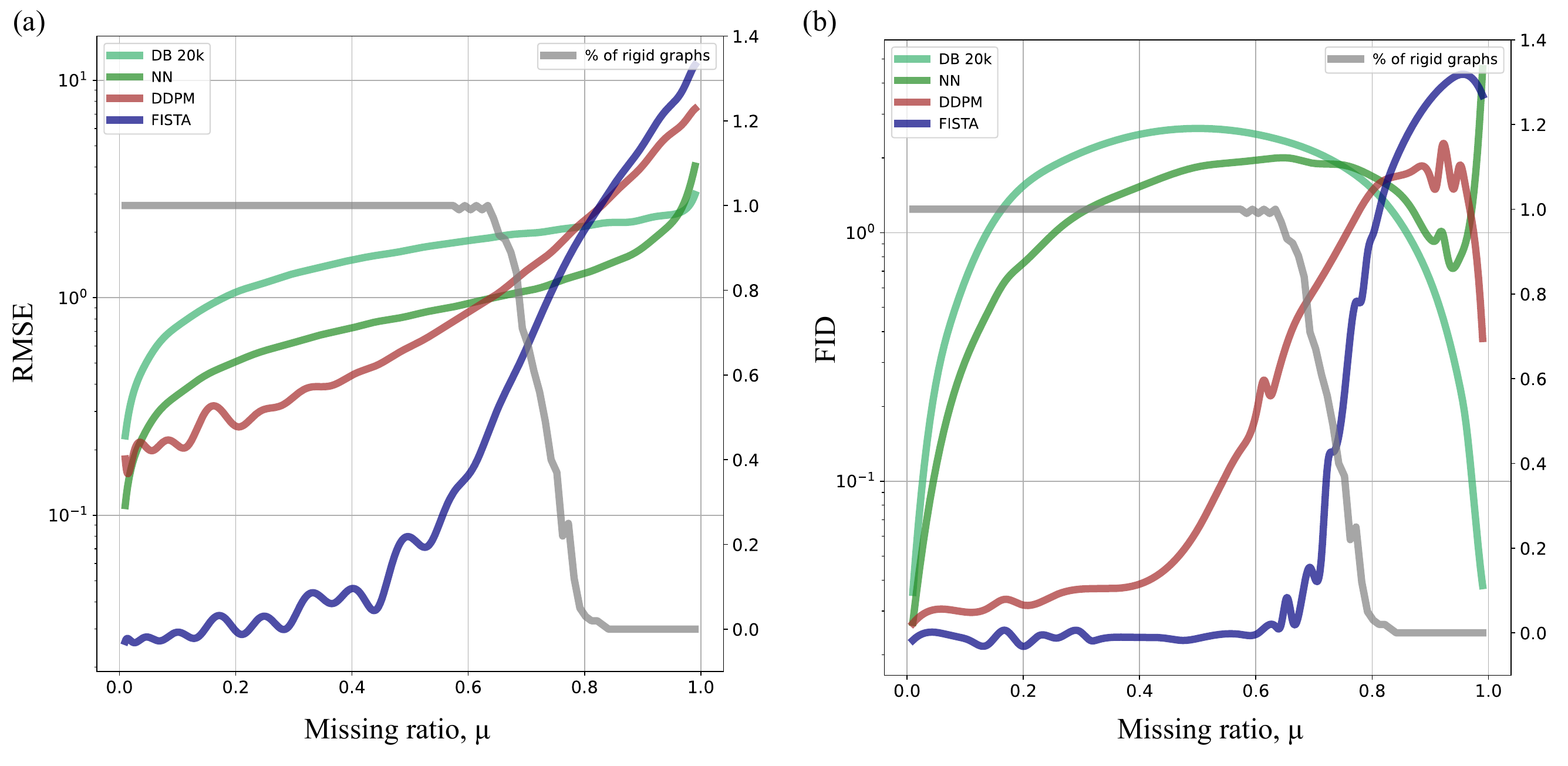}    
\caption{RMSE (a) and FID (b) plots as a function of missing ratio $\mu$ for different data imputation methods ($H=2/3$). The fraction of rigid graphs is shown in the second axis (grey).}    
\label{fig:figure_01}    
\end{figure}

\begin{table}[H]
\begin{center}
\small
    \begin{tabular}{|l|l|l|l|}
    \hline
    \textbf{Methods }       & $\frac{\sqrt{\sum_{i=1}^{5} \lambda^{2}_{i}}}{\sqrt{\sum_{i=1}^{64} \lambda^{2}_{i}}}$, $H=1/3$  & $\frac{\sqrt{\sum_{i=1}^{5} \lambda^{2}_{i}}}{\sqrt{\sum_{i=1}^{64} \lambda^{2}_{i}}}$, $H=1/2$ & $\frac{\sqrt{\sum_{i=1}^{5} \lambda^{2}_{i}}}{\sqrt{\sum_{i=1}^{64} \lambda^{2}_{i}}}$, $H=2/3$  \\ \hline
    Database search & $0.969 \pm 0.011$ & $0.982 \pm 0.008$ & $0.992 \pm 0.006$ \\
    Nearest Neighbour & $0.982 \pm 0.007$ & $0.993 \pm 0.003$ & $0.9972 \pm 0.0018$ \\
    DDPM & $\textbf{0.9977} \pm \textbf{0.0016}$ & $\textbf{0.99980} \pm \textbf{0.00018}$ & $\textbf{0.9997}\pm \textbf{0.0003}$\\
    \hline
    \textbf{Methods }       & $\frac{\sqrt{\sum_{i=1}^{5} |\lambda_{i}|}}{\sqrt{\sum_{i=1}^{64} |\lambda_{i}|}}$, $H=1/3$  & $\frac{\sqrt{\sum_{i=1}^{5} |\lambda_{i}|}}{\sqrt{\sum_{i=1}^{64} |\lambda_{i}|}}$, $H=1/2$ & $\frac{\sqrt{\sum_{i=1}^{5} |\lambda_{i}|}}{\sqrt{\sum_{i=1}^{64} |\lambda_{i}|}}$, $H=2/3$  \\ \hline
    Database search & $0.73 \pm 0.03$ & $0.77 \pm 0.03$ & $0.84 \pm 0.04$ \\
    Nearest Neighbour & $0.77 \pm 0.03$ & $0.83 \pm 0.03$ & $0.88 \pm 0.03$ \\
    DDPM & $\textbf{0.91} \pm \textbf{0.03}$ & $\textbf{0.974} \pm \textbf{0.0010}$ & $\textbf{0.968}\pm \textbf{0.014}$\\
    \hline
    \end{tabular}
\label{tab:res_comb}
\caption{The rank measures of the reconstructed distance matrices by different methods (database search, nearest neighbour, DDPM inpainting) at the missing ratio $\mu=0.5$ for three values of the Hurst parameter. The measures estimate the relative contribution of the first $r=5$ absolute values of the eigenvalues (or, squares of the eigenvalues) to the corresponding total.}
\end{center}
\end{table}

\begin{figure}[H]
\centering    
\includegraphics[width=\textwidth]{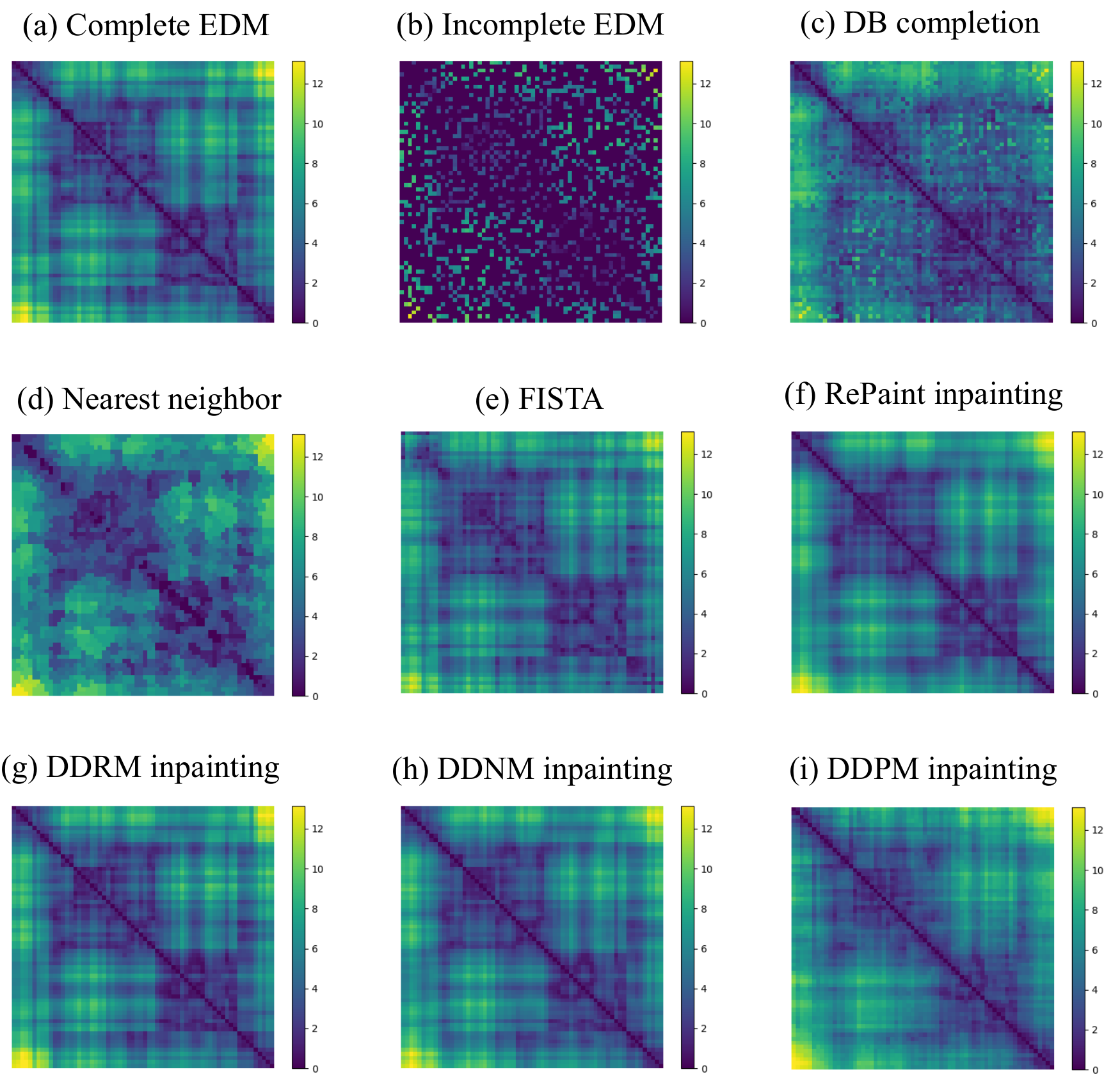}    
\caption{Original (a), incomplete (b) matrices and completions by different methods used in the paper, as indicated. The sparsity equals to $\mu=0.75$, for which no exact solution exists. Database completion is performed using the database size of $M=2*10^4$  trajectories. The Hurst parameter is $H=1/2$.}    
\label{fig:figure_01}    
\end{figure}

\begin{figure}[H]
\centering    
\includegraphics[width=\textwidth]{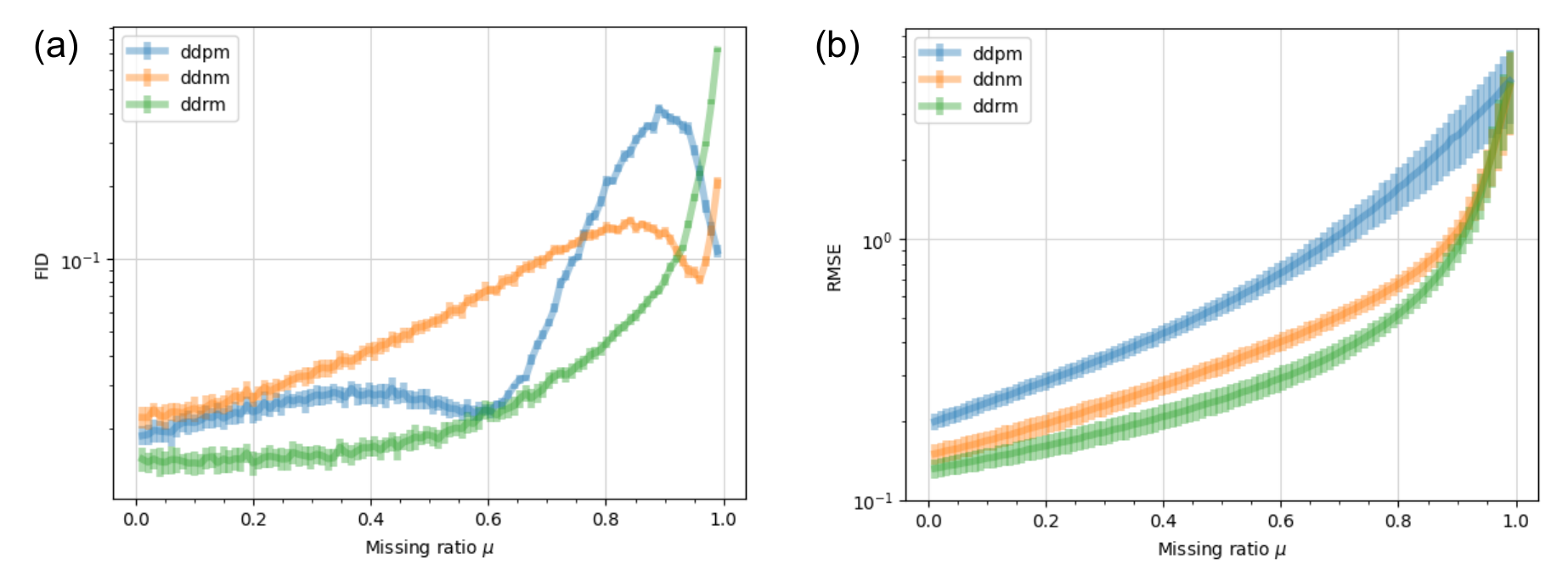}    
\caption{(a) FID and (b) RMSE for three diffusion-based inpainting methods. The metrics are computed as functions of sparsity $\mu$ of originally incomplete EDMs of fBm trajectories. The Hurst parameter of the corresponding fBm trajectories is $H=1/2$. The errors of RMSE are computed using a sample of 2000 inpainted distance matrices. The errors of FID for each $\mu$ are computed by randomly drawing (100 times) sub-samples with 90\% of matrices and computing the values of FID for each sub-sample; then the mean and the standard deviation of these values is taken.}
\label{fig:figure_01}    
\end{figure}

\begin{figure}[H]
\centering   
\includegraphics[width=0.7\textwidth]{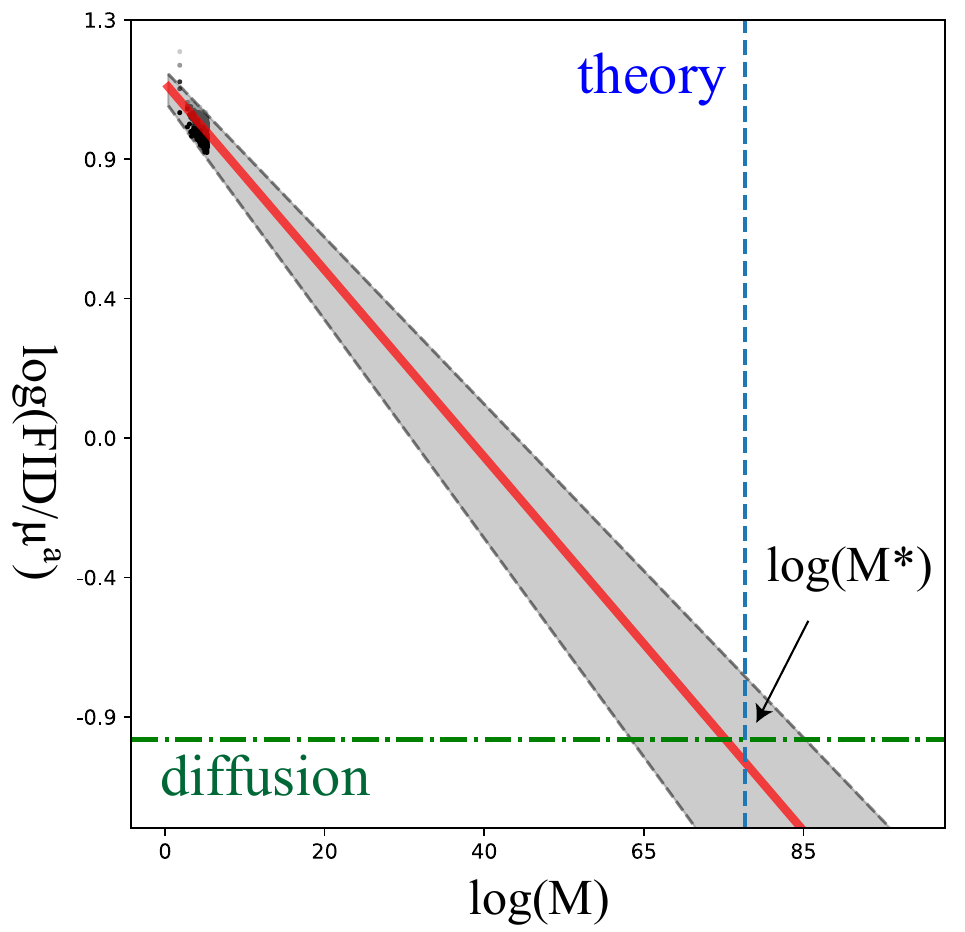}

  \caption{Log of FID from Fig. 3(c) optimally rescaled by $\mu^a$ with $a \approx 1.4$ and further extrapolated at larger database sizes. The red line is the optimal-slope line for the whole set of points at different sparsity; the grey strip provides the lower and the upper bound for the slope. The blue dashed line is the theoretical estimation of the effective database size $M^*$. The green dashed line is the mean FID of the diffusion-based inpainting for the considered range of $\mu$ further scaled by $\langle\mu \rangle^a$. The arrow indicates the effective database size $M^*$.}

\end{figure}

\begin{table}[H]  
\centering  
\begin{tabular}{|c|c|c|c|c|c|}  
\hline  
\textbf{Segment Index} & \textbf{Chromosome Index} & \textbf{n} & \textbf{Z} & \textbf{X} & \textbf{Y} \\  
\hline  
24181 & 373 & 1 & 8482 & 129943 & 64040 \\  
24182 & 373 & 2 & 8441 & 129908 & 64041 \\  
24183 & 373 & 3 & 8394 & 129955 & 64060 \\  
24184 & 373 & 4 & 8430 & 129788 & 64067 \\  
24185 & 373 & 5 & 8553 & 130238 & 64339 \\  
24186 & 373 & 6 & 8444 & 130165 & 64063 \\  
24187 & 373 & 7 & 8298 & 130144 & 64099 \\  
24188 & 373 & 8 & 8326 & 130297 & 64313 \\  
24189 & 373 & 9 & 8252 & 130136 & 64143 \\  
24190 & 373 & 10 & 8424 & 130273 & 64122 \\  
24191 & 373 & 11 & 8335 & 130272 & 64062 \\  
24192 & 373 & 12 & 8396 & 129771 & 63431 \\  
24193 & 373 & 13 & 8450 & 130445 & 63714 \\  
24194 & 373 & 14 & 8586 & 130527 & 63707 \\  
24195 & 373 & 15 & 8625 & 130672 & 63735 \\  
24196 & 373 & 16 & 8454 & 130620 & 63812 \\  
24197 & 373 & 17 & 8201 & 130362 & 64212 \\  
24198 & 373 & 18 & nan & nan & nan \\  
24199 & 373 & 19 & 8463 & 130814 & 64030 \\  
24200 & 373 & 20 & nan & nan & nan \\  
24201 & 373 & 21 & 8775 & 130133 & 63451 \\  
24202 & 373 & 22 & 8341 & 130432 & 64179 \\  
24203 & 373 & 23 & 8208 & 130258 & 64325 \\  
24204 & 373 & 24 & nan & nan & nan \\  
24205 & 373 & 25 & 8312 & 129874 & 64692 \\  
24206 & 373 & 26 & 8113 & 130130 & 64465 \\  
24207 & 373 & 27 & nan & nan & nan \\  
24208 & 373 & 28 & nan & nan & nan \\  
24209 & 373 & 29 & 8107 & 129700 & 64160 \\  
24210 & 373 & 30 & 7949 & 129708 & 64161 \\  
24211 & 373 & 31 & 7288 & 129982 & 63731 \\  
24212 & 373 & 32 & nan & nan & nan \\  
24213 & 373 & 33 & nan & nan & nan \\  
24214 & 373 & 34 & nan & nan & nan \\  
24215 & 373 & 35 & nan & nan & nan \\  
24216 & 373 & 36 & nan & nan & nan \\  
\hline  
\end{tabular}  
\caption{The 3D coordinates of 30kb segments (the center positions, in nm) on the region 28Mb-30Mb of chromosome 21 of the HCT116 cell 373 (part 1). Note that 10 nodes have been additionally dropped from the original data. (1,3,14,23,26,39,43,51,59,61)}  
\label{tab:my_label1}  
\end{table}

\begin{table}[H]  
\centering  
\begin{tabular}{|c|c|c|c|c|c|}  
\hline  
\textbf{Segment Index} & \textbf{Chromosome Index} & \textbf{n} & \textbf{Z} & \textbf{X} & \textbf{Y} \\  
\hline  
24217 & 373 & 37 & 8106 & 130225 & 64104 \\  
24218 & 373 & 38 & 7986 & 129517 & 64128 \\  
24219 & 373 & 39 & 7818 & 129537 & 64071 \\  
24220 & 373 & 40 & 7636 & 129434 & 64172 \\  
24221 & 373 & 41 & 7837 & 129253 & 64606 \\  
24222 & 373 & 42 & 7804 & 129358 & 64108 \\  
24223 & 373 & 43 & 7808 & 129358 & 64126 \\  
24224 & 373 & 44 & 7646 & 129311 & 64315 \\  
24225 & 373 & 45 & 7701 & 129525 & 64315 \\  
24226 & 373 & 46 & 7765 & 129656 & 64470 \\  
24227 & 373 & 47 & nan & nan & nan \\  
24228 & 373 & 48 & 8313 & 129185 & 64169 \\  
24229 & 373 & 49 & nan & nan & nan \\  
24230 & 373 & 50 & 7851 & 129993 & 64434 \\  
24231 & 373 & 51 & 8190 & 129829 & 64263 \\  
24232 & 373 & 52 & 8243 & 130238 & 64363 \\  
24233 & 373 & 53 & nan & nan & nan \\  
24234 & 373 & 54 & nan & nan & nan \\  
24235 & 373 & 55 & 8480 & 130410 & 64230 \\  
24236 & 373 & 56 & 8540 & 130605 & 64418 \\  
24237 & 373 & 57 & 8665 & 130560 & 64501 \\  
24238 & 373 & 58 & 8597 & 130403 & 64485 \\  
24239 & 373 & 59 & 8599 & 130559 & 64144 \\  
24240 & 373 & 60 & 8649 & 130703 & 64029 \\  
24241 & 373 & 61 & 8757 & 130610 & 64206 \\  
24242 & 373 & 62 & 8678 & 130609 & 64309 \\  
24243 & 373 & 63 & 8501 & 130637 & 64242 \\  
24244 & 373 & 64 & 8617 & 130734 & 64197 \\  
24245 & 373 & 65 & nan & nan & nan \\  
\hline  
\end{tabular}  
\caption{The 3D coordinates of 30kb segments (the center positions, in nm) on the region 28Mb-30Mb of chromosome 21 of the HCT116 cell 373 (part 2). Note that 10 nodes have been additionally dropped from the original data. (1,3,14,23,26,39,43,51,59,61)}  
\label{tab:my_label2}  
\end{table}  

\begin{figure}[H]
\centering   
\includegraphics[width=0.7\textwidth]{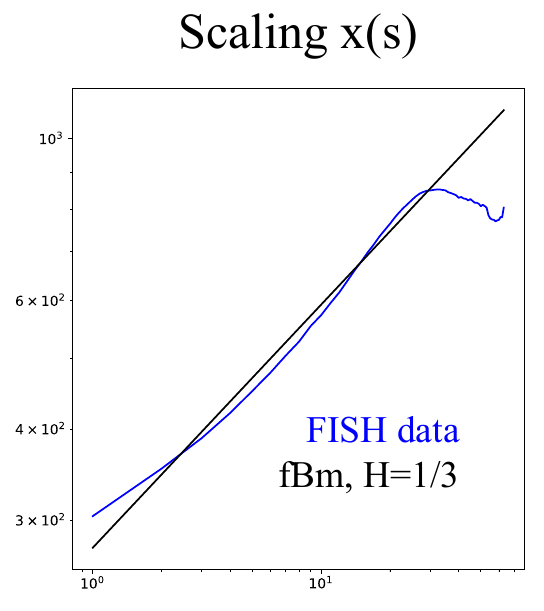}

  \caption{Scaling of the average spatial distance $x(s)$ between two loci separated by distance $s$ along chromosome. The spatial distance $x(s)$ is measured in nm, the chromosomal distance $s$ is measured in 30kb bins. The blue curve is computed from the FISH data by averaging along the $s$-th diagonal of the matrix corresponding to the cell shown in Figure 5. The black curve corresponds to the fBm trajectory with $H=1/3$, i.e. $\langle x_H^2(s) \rangle^{1/2} \sim s^{H}$, see Eq. 7. }

\end{figure}

\end{document}